\documentclass[]{fairmeta}
\pdfmapline{=optimistic < assets/Optimistic-pdftex.ttf <T1-WGL4.enc}
\microtypesetup{expansion=false}
\usepackage{amsmath}
\usepackage{amssymb}
\usepackage{mathtools}
\usepackage{colortbl}
\usepackage{array}
\usepackage{tabularx}
\usepackage{longtable}
\definecolor{jevvlaheader}{RGB}{213,229,241}
\definecolor{jevvladirect}{RGB}{232,243,252}
\definecolor{jevvlaours}{RGB}{220,239,222}
\definecolor{jevvladelta}{RGB}{239,247,241}
\definecolor{jevvlatotal}{RGB}{245,235,217}
\definecolor{jevvlasection}{RGB}{242,243,245}
\definecolor{jevvlapositive}{RGB}{45,105,76}
\definecolor{jevvlanegative}{RGB}{166,66,56}
\definecolor{jevvlagray}{RGB}{105,105,105}
\newcolumntype{L}{>{\raggedright\arraybackslash}X}
\newcolumntype{J}{>{\columncolor{jevvladirect}}r}
\newcolumntype{P}{>{\columncolor{jevvlaours}}r}
\newcolumntype{K}{>{\columncolor{jevvladelta}}r}
\newcommand{\jevvlagain}[1]{\textcolor{jevvlapositive}{\bfseries #1}}
\newcommand{\jevvlaloss}[1]{\textcolor{jevvlanegative}{#1}}
\newcommand{\jevvlaneutral}[1]{\textcolor{jevvlagray}{#1}}
\newcommand{\jevvlabest}[1]{\textcolor{jevvlapositive}{\bfseries #1}}

\title{JevVLA: Jev-Inspired Energy-Based Adaptation for Vision-Language-Action Policies}
\hypersetup{pdftitle={JevVLA: Jev-Inspired Energy-Based Adaptation for Vision-Language-Action Policies},pdfauthor={Wei Wang, Wenqiao Zhang, Zheqi Lv, Zheng Dong, Yingjie Niu, Jiaqi Zhu, Jun Xiao, Chao Li, Yueting Zhuang}}

\author[1]{Wei Wang}
\author[1,\ddagger]{Wenqiao Zhang}
\author[2]{Zheqi Lv}
\author[3]{Zheng Dong}
\author[4]{Yingjie Niu}
\author[5]{Jiaqi Zhu}
\author[1]{Jun Xiao}
\author[4]{Chao Li}
\author[1]{Yueting Zhuang}

\affiliation[1]{College of Computer Science and Technology, Zhejiang University}
\affiliation[2]{Cornell University}
\affiliation[3]{Universal Ubiquitous AI Co., Ltd.}
\affiliation[4]{Hangzhou DEEP Robotics Technology Co., Ltd.}
\affiliation[5]{National University of Singapore}
\contribution[\ddagger]{Corresponding author}

\abstract{Vision--language--action (VLA) policies can select among discrete candidates
for pick-and-place manipulation or generate continuous action chunks for
motion control. Improving both forms of decision making through a common
mechanism while retaining the capabilities of a pretrained policy is an
important adaptation problem. Inspired by Jev's structured, probabilistic
decision interface, we propose \textbf{JevVLA}, a lightweight energy-based
adaptation framework for VLA policies. JevVLA instantiates this inspiration
by modeling the relative quality of action proposals: with the base policy
frozen, it learns a differentiable scalar energy conditioned on observations,
task context, and actions, using execution feedback, expert corrections, or
demonstration-derived preferences. JevVLA uses this single learned action
preference function in two ways: its values compare discrete candidates,
while its action gradients correct continuous actions or guide their
generation. Candidate selection and continuous control thus share one
trainable decision core.
Each benchmark uses independently trained weights and fixed input/output
encodings, while retaining the same core network with approximately 80,000
trainable parameters. On the held-out test sets, JevVLA improves task success over the respective
matched base-policy controls by 1.28, 2.31, 1.60, 1.52, and 1.28 percentage points on MetaWorld
MT50, VIMA, LanguageTable, LIBERO Plus, and CLIPort, respectively. These
results support the feasibility of a lightweight, Jev-inspired energy model
for adapting both discrete action selection and continuous action guidance.
The JevVLA implementation will be open-sourced at
\url{https://github.com/ZJU4EmbodiedAI/JevVLA}.
}

\begin{document}
\maketitle

\noindent\begin{minipage}{\linewidth}
\captionsetup{type=figure,skip=5pt}
\centering
\includegraphics[width=\linewidth]{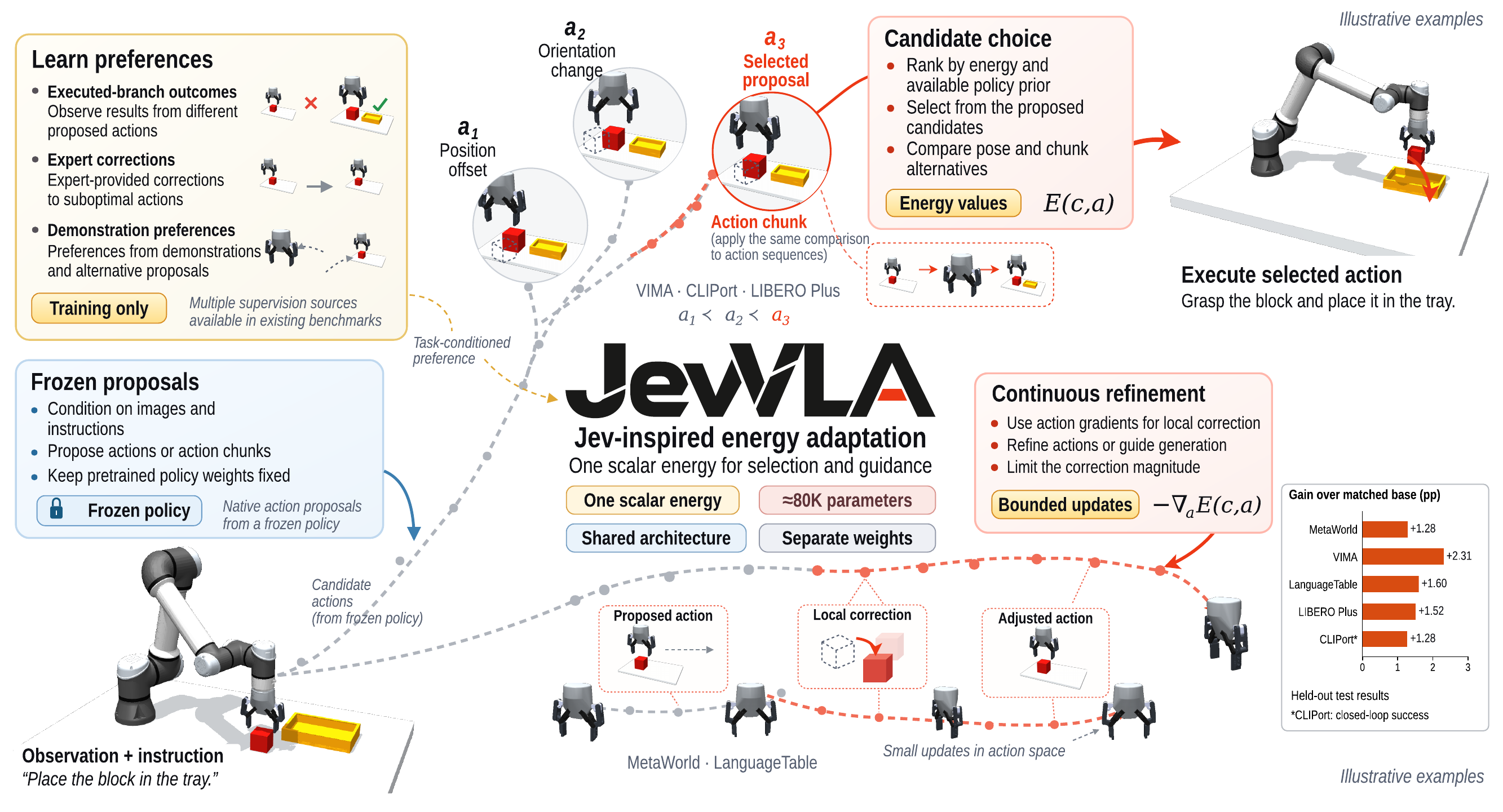}\par
\caption{\textbf{Overview of JevVLA.}
A lightweight Jev-inspired scalar energy adapts frozen robot policies:
its values compare action candidates, while its action gradients provide
bounded corrections or guide generation. Benchmarks share the core
architecture and use independently trained weights. Robot scenes are local simulator renders used as illustrations, not
JevVLA rollout frames. Bars show the test-set gains in Table~\ref{tab:jevvla-overall}.
CLIPort reports strict terminal success from closed-loop execution.
Direct evaluates the same frozen base policy without JevVLA.}
\label{fig:jevvla-teaser}
\end{minipage}\par

\section{Introduction}
\label{sec:introduction}

Vision--language--action (VLA) models connect visual observations and task
instructions to robot actions. By combining vision--language pretraining
with robot demonstrations, these models learn control policies that serve
multiple manipulation tasks, making the reuse and adaptation of existing
models an important direction in robot learning~\cite{pmlr-v270-kim25c,pmlr-v305-black25a}.
However, producing a semantically plausible action does not ensure that an
operation will succeed in a particular scene: small geometric errors can
change the execution outcome~\cite{ijcai2026p569}.
Further fine-tuning a pretrained policy, particularly through
full-parameter updates, often entails substantial computational costs and
additional task-specific data~\cite{pmlr-v270-kim25c,yang2026_260702092}.
This motivates a practical question: how can lightweight adaptation improve
action decisions while retaining the action-generation capabilities of the
base policy?

This goal must accommodate differences in the action interfaces of robot
policies. Policies that predict pick-and-place actions can be adapted by
constructing a finite set of candidates and comparing their quality.
Policies that control robots through continuous action sequences can also
be adapted through corrections applied after action prediction or during
iterative generation~\cite{pmlr-v202-jiang23b,pmlr-v164-shridhar22a,Chi-RSS-23,black2024_241024164}.
Both approaches require an assessment of relative action quality, but they
place different demands on the evaluator: candidate selection requires
comparable scores across actions, whereas continuous guidance requires
useful local correction directions in action space. We investigate how a
single lightweight evaluation architecture can support both requirements
across base policies with different native action representations.

Prior work has shown that scalar energies can represent implicit action
policies and support robot control through action optimization~\cite{pmlr-v164-florence22a}.
For policy adaptation, learned value or performance estimates can select
among sampled action chunks or guide continuous action generation through
gradients~\cite{ijcai2026p569,yang2026_260702092,wang2026_260310980}.
These advances establish the utility of action evaluation for improving
base policies. We study how to reuse the
function values and action gradients of a common scalar core to organize
candidate comparison and continuous correction within one adaptation
framework, while keeping the base policies frozen.

Jev provides an inspiration for this design through its structured,
probabilistic decision interface. Its public interface combines a state
with predefined options or scoring questions, returning structured results
and probability distributions that software can use for selection, ranking,
and branching~\cite{almeida2026jev,typesafe2026jevintro}.
This perspective motivates modeling action proposal and action evaluation
as separate components: the base policy proposes actions, and an adapter
represents preferences among them given the current observation and task
context. JevVLA draws on Jev's public decision-interface concepts; its robotic
action energy, training objectives, and gradient-guidance mechanism are
developed independently. We extend explicit comparison among candidates to
continuous action space, using a differentiable scalar function both to
support selection and to provide local correction directions.

We propose \textbf{JevVLA}, a lightweight, Jev-inspired energy-based adaptation
framework (Figure~\ref{fig:jevvla-teaser}). JevVLA takes observations, task context, and actions as inputs and
uses a conditional quadratic term together with a scalar residual network
to model action energy. Depending on the supervision available for each
benchmark, it converts executed-branch outcomes, expert corrections, or
demonstration-derived action preferences into ranking or gradient-learning
signals. At inference time, JevVLA compares energies over a finite candidate
set and incorporates the base-policy prior when available. For continuous
actions, it differentiates the same energy with respect to the action and
applies bounded corrections or guides action generation. The architectural
contribution is one trainable preference function for these two decision
interfaces: its value supports choice, and its derivative supplies the
continuous update. Each benchmark has fixed input/output encodings and
independently trained weights, while the energy core remains identical,
with \textbf{79,817 trainable parameters} per adapter.

We evaluate this design on five robot manipulation benchmarks: MetaWorld
MT50, VIMA, LanguageTable, LIBERO Plus, and
CLIPort~\cite{pmlr-v100-yu20a,pmlr-v202-jiang23b,lynch2022_221006407,Fei_2026_CVPR,pmlr-v164-shridhar22a}.
These evaluations cover both candidate selection and continuous control.
After validation-set selection, JevVLA improves held-out test success over
the respective matched base-policy controls by
\textbf{1.28, 2.31, 1.60, 1.52, and 1.28 percentage points}.
These results provide empirical support for applying the same lightweight
energy architecture across different robot-policy interfaces.

Our contributions are threefold:
\begin{itemize}
    \item \textbf{One action preference core for two VLA interfaces.}
    Values of a conditional scalar energy rank discrete proposals, while
    gradients of that energy correct continuous actions and guide their
    generation. The same trainable architecture supports both uses across
    benchmarks, with independently learned weights.
    \item \textbf{Jev-inspired structured action decisions.}
    We make alternatives and their relative preferences explicit for
    frozen robot policies, connecting action proposals to a probabilistic
    comparison and executable selection rule.
    \item \textbf{Evaluation across five robot manipulation benchmarks.}
    Independently trained adapters with approximately 80,000 parameters
    improve task success over matched base-policy controls in the specified
    held-out test settings, providing cross-interface evidence for
    the common architecture.
\end{itemize}

\section{Related Work}
\label{sec:related_work}

\subsection{Vision--language--action models}

Vision--language robot policies connect images, task instructions, and actions,
bringing semantic knowledge and interaction experience to manipulation.
Early robot Transformers and subsequent VLA models demonstrate the value of
multitask learning and vision--language transfer, while open models provide
reusable weights and adaptation
pipelines~\cite{Brohan-RSS-23,pmlr-v229-zitkovich23a,pmlr-v270-kim25c}.
Cross-embodiment pretraining, flow-based action generation, and heterogeneous
co-training expand policy coverage; compact vision--language backbones and
asynchronous execution also explore training and deployment
efficiency~\cite{Ghosh-RSS-24,black2024_241024164,pmlr-v305-black25a,shukor2025_250601844}.

These policies expose different action interfaces. Semantic--spatial
decomposition, multimodal prompting, and real-time language-conditioned
control support pick-and-place decisions or ongoing motion
control~\cite{pmlr-v164-shridhar22a,pmlr-v202-jiang23b,lynch2022_221006407}.
Action chunking and diffusion-based sequences extend individual predictions
to short-horizon generation~\cite{Zhao-RSS-23,Chi-RSS-23}.
For downstream adaptation, prior work improves action decoding and fine-tuning
objectives or learns a new sampling distribution over the latent-noise space
of a frozen policy~\cite{KimM1-RSS-25,pmlr-v305-wagenmaker25a}.
Reusing an adapter across policies requires handling these action-interface
differences: finite-candidate comparison and continuous correction require
different operations, and guidance tied to a particular generator has a
corresponding scope of applicability. To accommodate these differences,
JevVLA extracts action evaluation into a separate conditional scalar module,
connecting policies through fixed encodings and using the same core
structure for candidate selection and continuous guidance.

\subsection{Robot preference learning}

Robot preference learning expresses difficult-to-specify objectives through
the relative quality of behaviors. Iterative trajectory improvement and
joint modeling of demonstrations, corrections, and preferences show that
supervision can come from alternatives that improve on current
behavior~\cite{jain2013_13066294,mehta2022_220703395}.
Reward learning from human comparisons converts segment preferences into
scalar objectives, while experience relabeling, unsupervised pretraining,
and data augmentation improve feedback
utilization~\cite{NIPS2017_d5e2c0ad,pmlr-v139-lee21i,park2022_220310050}.

Interactive imitation learning queries experts or records human interventions
at states visited by the policy, using data aggregation, sample reweighting,
and intervention budgets to improve subsequent
behavior~\cite{pmlr-v15-ross11a,liu2022_221108416,pmlr-v164-hoque22a}.
Work on heterogeneous demonstrations and robotic action preferences learns
behavior-quality evaluators or directly updates policies to favor corrected
actions over unsuccessful
ones~\cite{pmlr-v229-kuhar23a,NEURIPS2025_347110fb}.
Trajectory quality, terminal success, and local expert corrections
provide feedback at different time scales.
Mismatched preference and optimization scales affect credit
assignment~\cite{taranovic2026_260611982}.
JevVLA accordingly constructs ranking or gradient supervision from each data
source, preserving the distinct meanings of branch outcomes, corrected
actions, and demonstration distances. These signals train a separate energy
adapter with a common core architecture, while the available supervision
determines the loss used in each setting.

\subsection{Energy-based action selection}

Energy models represent relative action preferences or unnormalized
distributions with conditional scalar functions. Maximum-entropy
reinforcement learning connects action values to energy-based policies;
implicit imitation learning learns complex action distributions by comparing
expert actions with negative samples, with subsequent work analyzing ranking
objectives and negative
sampling~\cite{pmlr-v70-haarnoja17a,pmlr-v164-florence22a,singh2023_230905803}.
Implicit kinematic policies and compositional scene planning further show
that energy representations can connect action coordinates or combine spatial
relations, acting on control representations and target layouts,
respectively~\cite{ganapathi2022_220301983,gkanatsios2023_230414391}.

Visual action-value learning, finite-candidate maximization over behavior
models, and value-based reweighting of diffusion candidates provide different
implementations of proposal followed by
evaluation~\cite{pmlr-v87-kalashnikov18a,pmlr-v139-ghasemipour21a,hansenestruch2023_230410573}.
This approach also supports few-shot VLA adaptation through geometrically
conditioned evaluation of similar action chunks with different execution
outcomes~\cite{ijcai2026p569}.
However, finite-candidate selection remains limited by the proposal
distribution and candidate set. Correctly ranking observed actions does not
necessarily provide reliable local gradients between them. Although existing
energy policies already employ sampling and optimization, directly learning
a control distribution differs from adapting a frozen policy that retains
its native interface. JevVLA uses ranking and gradient supervision in its
respective training settings and assigns both functions to the same energy
architecture, studying its use as an adapter for frozen policies.

\subsection{Inference-time action guidance}

Score matching and denoising connect probabilistic models to input gradients,
while diffusion and flow matching provide different iterative paths for
generating complex
samples~\cite{JMLR:v6:hyvarinen05a,Vincent_2011,song2021scorebased,NEURIPS2020_4c5bcfec,lipman2023flow}.
Classifier gradients and combinations of conditional and unconditional
predictions modify sampling preferences in conditional generation; trajectory
diffusion brings generative priors together with test-time reward guidance
for decision making~\cite{NEURIPS2021_49ad23d1,ho2022_220712598,pmlr-v162-janner22a}.
These studies provide technical foundations for steering generation with
external objectives.

For robot policies, action-chunk critics provide value gradients to frozen
flow policies, while success/failure predictors and retrieved execution
traces supply other correction
signals~\cite{yang2026_260702092,zhou2026_260611087,wang2026_260310980,zhang2026_260920646}.
Other approaches derive task conditions from vision--language rewards,
transferable contact geometry, or motion
primitives~\cite{liu2026_260203973,wu2024_241203142,li2026_260707076}.
For nondifferentiable costs, sampling and path-integral methods also offer
optimization without action gradients~\cite{park2026_260721661}.
The guidance signals and interfaces in these methods vary with value
estimation, supervision, retrieval, and generator structure; different
generative processes also require appropriate guidance
approximations~\cite{pmlr-v267-feng25s}.
JevVLA focuses on the architectural question of how a model used for continuous
correction can also evaluate action candidates from nongenerative policies.
We use the values of a differentiable scalar core for candidate comparison
and its action gradients for bounded continuous correction, determining
gradient insertion and scaling from the base-policy interface.

\subsection{Structured decisions with Jev}

Jev's public interface combines states with predefined typed questions and
returns structured decisions directly usable by
software~\cite{almeida2026jev,typesafe2026jevintro}.
Its choice interface returns candidates and their probabilities, while its
scoring interface evaluates a predefined scale, connecting model judgments
to subsequent ranking and branching
operations~\cite{typesafe2026choice,typesafe2026score}.
This explicit representation of candidate preferences motivates separating
action proposal from action evaluation.

We draw on Jev's structured probabilistic decision perspective. Applying it
to robot control requires action representations, training objectives for
execution feedback, and continuous guidance mechanisms, which Jev's public
interface does not specify. Building on energy policies, preference learning,
and generative guidance, JevVLA independently implements a robot-specific
conditional energy adapter that connects discrete candidate evaluation with
continuous correction. The unification studied here is reuse of the same
core architecture across benchmarks, with weights trained separately for
each benchmark; reproducing Jev's internal model is outside this scope.

\section{Method}
\label{sec:method}

\begin{figure}[!htbp]
\centering
\includegraphics[width=\linewidth]{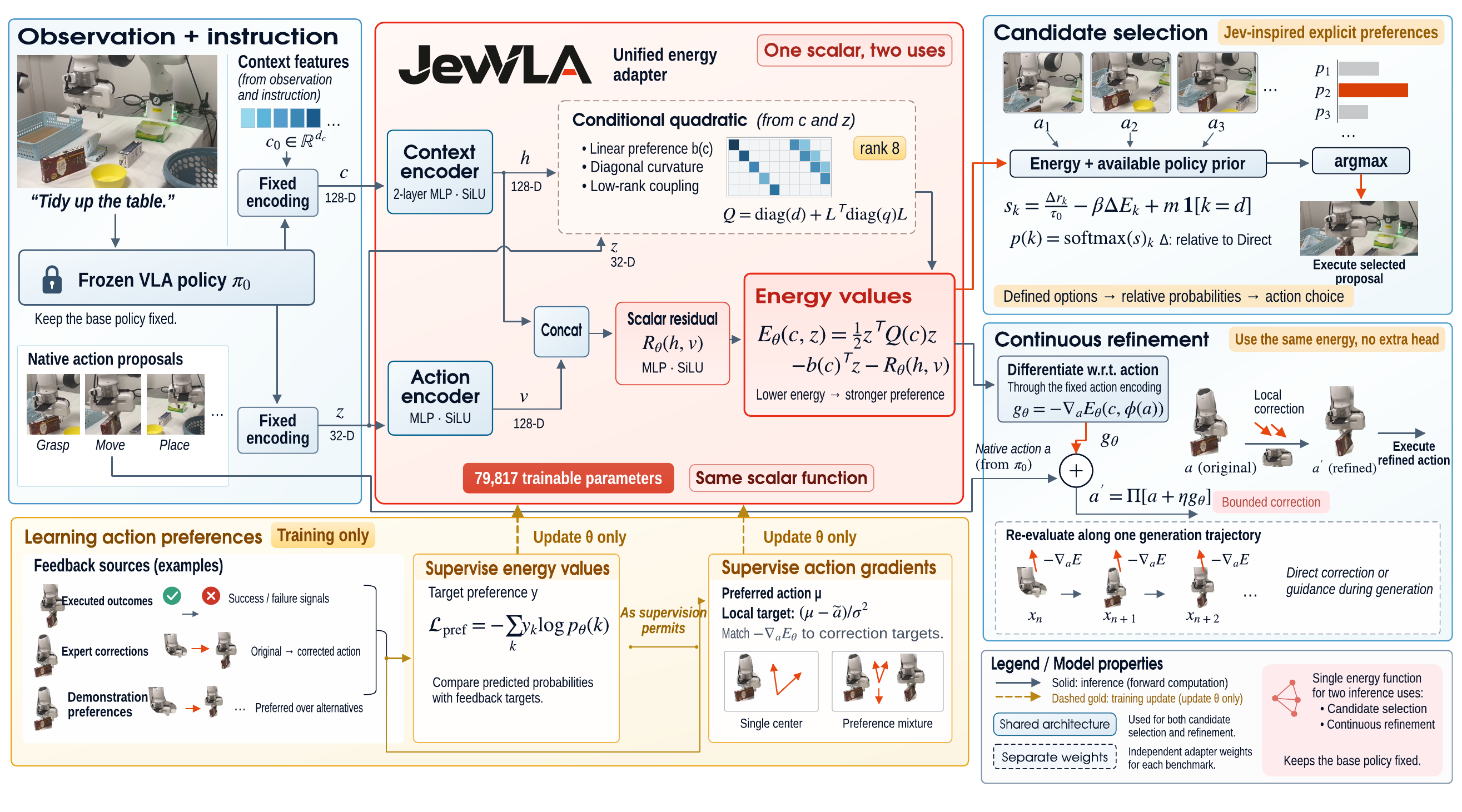}
\caption{\textbf{JevVLA architecture.} Fixed encodings connect observations and
native action proposals from a frozen policy to a conditional energy core.
The core combines a quadratic potential with a nonlinear scalar residual.
Its values rank candidates, optionally with the policy prior, while its
action gradients provide bounded corrections or guide iterative generation.
Execution feedback, expert corrections, or demonstration preferences
supervise energy values and action gradients as available; training updates
only the adapter. Benchmarks share the 79,817-parameter core architecture
and learn separate weights. Robot images illustrate the computation.}
\label{fig:jevvla-architecture}
\end{figure}

JevVLA adapts a frozen robot policy by learning how to evaluate its actions.
The shared decision mechanism is a conditional scalar energy: its values
choose among discrete proposals, and its action derivatives steer continuous
predictions. One trainable preference architecture therefore serves both
action interfaces, with independent weights and fixed encodings for each
benchmark.
We first describe the connection to Jev's decision interface, then specify
the shared energy architecture, its supervision, and its use at inference.
Figure~\ref{fig:jevvla-architecture} shows the computation from frozen-policy
proposals to energy-based decisions and the feedback used to train the core.

\subsection{Structured action preferences}
\label{sec:method:preferences}

Let $o$ denote the current observation, including available visual and robot
state inputs, and let $\ell$ denote the task instruction or multimodal prompt.
A frozen policy $\pi_0$ produces a native action $a$, which may represent a
pick-and-place command, a displacement, or an action chunk. A fixed context
encoder $\chi_b$ and action encoder $\phi_b$ connect benchmark $b$ to the
adapter:
\begin{equation}
    c=\chi_b(o,\ell;\pi_0)\in\mathbb{R}^{128},\qquad
    z=\phi_b(a)\in\mathbb{R}^{32},\qquad E_\theta(c,z)\in\mathbb{R}.
    \label{eq:method:interface}
\end{equation}
Context encodings reuse frozen policy or perception features and can include
generation time, control progress, or current proposal summaries. Their
parameters are fixed during energy training; fitted projections use training
data only. We adopt the convention that lower energy indicates a preferred
action. The energy is a learned preference potential, not physical work.

Jev's public \emph{Choice} primitive evaluates predefined alternatives and
returns a distribution over them; its \emph{Score} primitive evaluates
ordered, descriptive levels~\cite{typesafe2026choice,typesafe2026score}.
The relevant design principle is to make the decision alternatives and their
relative preferences explicit. JevVLA instantiates this principle with robot
actions: the base policy supplies proposals, the observation supplies decision
context, and an energy-induced distribution supplies preferences that control
action selection. A candidate can be viewed as an outgoing action branch
from the current observation.

Our connection to Jev is at this decision-interface level. The public Jev
materials do not provide a complete model and training recipe for
reproduction~\cite{almeida2026jev,typesafe2026jevintro}.
The robot action encodings, quadratic energy, feedback objectives, and
continuous guidance below are our own implementation.

\subsection{Conditional energy core}
\label{sec:method:core}

We parameterize the energy as a conditional quadratic potential with a
nonlinear scalar residual:
\begin{align}
    h &= f_\theta^{c}(c), \qquad v=f_\theta^{a}(z), \nonumber\\
    Q_\theta(c)&=\operatorname{diag}(d_\theta(h))
       +L^\top\operatorname{diag}(q_\theta(h))L, \label{eq:method:precision}\\
    E_\theta(c,z)&=\tfrac12 z^\top Q_\theta(c)z
       -b_\theta(h)^\top z-R_\theta(h,v).
       \label{eq:method:energy}
\end{align}
Here, $d_\theta(h)\in\mathbb{R}^{32}$ and
$q_\theta(h)\in\mathbb{R}^{8}$ are positive softplus outputs, and
$L\in\mathbb{R}^{8\times32}$ is learned. The diagonal term provides
coordinate-specific curvature; the low-rank term couples action coordinates,
including different time steps when $z$ encodes a chunk. The linear term
places the preferred region according to context. The residual captures
preferences that a quadratic potential cannot express.

The context network has two width-128 linear layers with SiLU activations.
The action network has one width-128 SiLU layer. Their concatenated features
feed a width-128 SiLU layer and a scalar output. Three linear projections of
$h$ produce $b_\theta$, $d_\theta$, and $q_\theta$. This structure contains
\textbf{79,817 trainable parameters}, with the same network structure
on all five benchmarks and independent weights for each benchmark.

Differentiating the energy with respect to the encoded action gives:
\begin{equation}
    -\nabla_z E_\theta(c,z)
    =b_\theta(h)-Q_\theta(c)z+\nabla_z R_\theta(h,f_\theta^{a}(z)).
    \label{eq:method:feature_gradient}
\end{equation}
The quadratic part supplies a context-dependent restoring direction, while
the residual supplies a learned nonlinear adjustment. For a differentiable action encoder, the native-action
correction field is obtained by the chain rule:
\begin{equation}
    g_\theta(c,a)=-\nabla_a E_\theta(c,\phi_b(a))
       =-J_{\phi_b}(a)^\top\nabla_z E_\theta(c,z).
    \label{eq:method:native_gradient}
\end{equation}
Ranking and correction thus share the same scalar parameters.

\subsection{Learning action preferences}
\label{sec:method:training}

A useful adapter must assign sensible values to competing actions and,
when continuous correction is required, provide useful derivatives around
them. We supervise the scalar through preference comparisons, through
action-gradient targets, or through both, according to the available feedback.

\paragraph{Probabilistic comparison.}
For an anchor observation, let $\mathcal{A}=\{a_k\}_{k=1}^{K}$ be the candidate
set and $d$ the unmodified base-policy proposal, referred to as Direct.
Write $E_k=E_\theta(c,\phi_b(a_k))$. Let $r_k$ be a native proposal score
when available, or zero otherwise. We define
\begin{align}
    s_k &=\frac{r_k-r_d}{\tau_0}
          -\beta(E_k-E_d)+m\,\mathbf{1}[k=d], \label{eq:method:logits}\\
    p_\theta(k\mid c,\mathcal{A})
        &=\frac{\exp(s_k)}{\sum_{j\in\mathcal{V}}\exp(s_j)},
          \quad k\in\mathcal{V}, \label{eq:method:probability}
\end{align}
where $\mathcal{V}$ is the valid candidate set, $\tau_0>0$ scales the native
score, $\beta\geq0$ controls the learned preference, and $m\geq0$ favors
retaining Direct. Invalid candidates receive zero probability. Subtracting
the Direct reference removes a common offset without changing relative
preferences.

Given target preferences $y\in\Delta^{K-1}$, we minimize
\begin{equation}
    \mathcal{L}_{\mathrm{pref}}
       =-\mathbb{E}\!\left[\sum_{k\in\mathcal{V}}y_k
                       \log p_\theta(k\mid c,\mathcal{A})\right].
    \label{eq:method:ce}
\end{equation}
Preference training uses unit energy gain and zero Direct margin; inference
can adjust these global constants. This formulation follows relative
feedback learning and energy-based action
evaluation~\cite{NIPS2017_d5e2c0ad,pmlr-v164-florence22a}.

For executed candidates with binary terminal outcomes, a pairwise variant
compares every successful candidate $i$ with every failed candidate $j$:
\begin{equation}
  \mathcal L_{\mathrm{pair}}
  =\mathbb E_{x:\lvert\mathcal P_x\rvert>0}
   \frac{1}{\lvert\mathcal P_x\rvert}
   \sum_{(i,j)\in\mathcal P_x}
   \operatorname{softplus}\!\left[-(s_i-s_j)\right],
  \quad \mathcal P_x=\{(i,j):y_i^{\rm term}>y_j^{\rm term}\}.
  \label{eq:method:pairwise}
\end{equation}
Each informative observation receives equal weight; observations whose
candidates all share the same terminal outcome have no pairwise term.
The CLIPort Full configuration uses cross-entropy in
Eq.~\eqref{eq:method:ce}. The low-rank ablation in
Table~\ref{tab:jevvla-ablations} uses the pairwise variant, so its
difference from Full includes a change of training objective.

We construct $y$ from the evidence available at the anchor. Executed branch
success can identify preferred alternatives; future expert corrections can
refine their ordering; proximity to demonstration actions provides an
offline preference when branch outcomes are unavailable. For a nonnegative
action discrepancy $D_k$, a soft target takes the form
$y_k\propto\exp(-D_k/\tau_y)$ on the eligible alternatives, with $\tau_y$
determined from fit data. Observed success can restrict that support before
the discrepancy weights are normalized. Demonstration labels measure
action proximity, and expert information supplies training targets.
Deployment uses the candidates provided by the base policy.

\paragraph{Learning the action field.}
For continuous adaptation, we train
$g_\theta(c,a)=-\nabla_a E_\theta(c,\phi_b(a))$ using correction targets.
This draws on score matching and denoising
objectives~\cite{JMLR:v6:hyvarinen05a,Vincent_2011}.
A local Gaussian centered at a preferred action $\mu$ defines the target
\begin{equation}
    g^*(\widetilde a;\mu,\sigma)
       =\frac{\mu-\widetilde a}{\sigma^2},\qquad
    \mathcal{L}_{\mathrm{grad}}
       =\mathbb{E}_{w}\!
          \left[\left\|M\bigl(g_\theta(c,\widetilde a)
                         -g^*(\widetilde a)\bigr)\right\|_2^2\right].
    \label{eq:method:gradient_loss}
\end{equation}
Here, $M$ selects the supervised action coordinates and $\mathbb{E}_{w}$
denotes the weighted average over training states and branches. The
center can be an expert action, a branch action sampled according to its
preference, or an executed action shifted toward a queried expert:
$\mu=a+\alpha(a^*-a)$. This construction converts an available correction
into a local direction field with an explicit noise scale.

When $\widetilde a=\mu+\sigma\epsilon$ and
$\epsilon\sim\mathcal{N}(0,I)$, the corresponding noise-scaled objective is
\begin{equation}
    \mathcal{L}_{\mathrm{DSM}}
       =\mathbb{E}\!\left[
          \frac{1}{d_a}
          \left\|\sigma g_\theta(c,\mu+\sigma\epsilon)
                          +\epsilon\right\|_2^2\right],
    \label{eq:method:dsm}
\end{equation}
where $d_a$ is the action dimension. Training on a range of noise scales
provides supervision in neighborhoods of preferred actions.
When several branches remain plausible, a preference-weighted Gaussian
mixture gives a target that preserves their alternatives:
\begin{equation}
    g_{\mathrm{mix}}^*(\widetilde a)
      =\sum_k\omega_k(\widetilde a)\frac{\mu_k-\widetilde a}{\sigma^2},
    \qquad
    \omega_k(\widetilde a)
      =\frac{\rho_k\exp\!\left(-\frac{\|\widetilde a-\mu_k\|^2}{2\sigma^2}\right)}
             {\sum_j\rho_j\exp\!\left(-\frac{\|\widetilde a-\mu_j\|^2}{2\sigma^2}\right)}.
    \label{eq:method:mixture}
\end{equation}
The masses $\rho_k$ come from recorded branch quality and expert alignment.
These posterior weights express which preferred branch is locally relevant
to the current action hypothesis. The mixture target and the single-branch
target are constructed analytically from the supervision data.

Gradient targets can be evaluated at perturbed actions or at recorded
intermediate states of the frozen action generator. Expert actions at
later visited environment states also extend supervision beyond initial
anchors. The actual objective uses the available preference and gradient
terms with fixed weights. All parameter updates act on the scalar energy,
including when the loss differentiates its action gradient; the base policy
remains frozen.

\subsection{Acting on the learned preference}
\label{sec:method:inference}

\paragraph{Selecting a proposal.}
For finite candidates, JevVLA evaluates their energies and executes
$a_{k^*}$, where $k^*=\arg\max_{k\in\mathcal{V}}s_k$.
For two valid candidates,
\begin{equation}
    \log\frac{p_\theta(k)}{p_\theta(j)}
    =\frac{r_k-r_j}{\tau_0}-\beta(E_k-E_j)
       +m\bigl(\mathbf{1}[k=d]-\mathbf{1}[j=d]\bigr).
    \label{eq:method:odds}
\end{equation}
The energy difference therefore changes the relative weight of concrete
action branches. This is the direct realization of Jev-inspired explicit
candidate preference. With zero native scores, a challenger replaces Direct
only if its energy advantage exceeds the margin. Execution uses the
maximum-score candidate, so computing the softmax normalization is optional.

\paragraph{Refining an action.}
For a continuous prediction $a_0$, the same preference potential supplies
a local update:
\begin{equation}
    a'=\Pi_{\mathcal{B}_b}
       \left(a_0+\eta g_\theta(c,a_0)\right),
    \label{eq:method:correction}
\end{equation}
where $\eta\geq0$ is a global gain and $\Pi_{\mathcal{B}_b}$ applies the
native action bounds. Before projection, for fixed $c$ and a sufficiently
small step, the first-order energy change is
\begin{equation}
    E_\theta(c,\phi_b(a_0+\eta g_\theta))
       =E_\theta(c,\phi_b(a_0))
        -\eta\|g_\theta(c,a_0)\|_2^2+O(\eta^2).
    \label{eq:method:descent}
\end{equation}
The negative energy gradient moves the action toward lower energy in this
local first-order approximation, turning scalar preference into action correction.

For iterative action generation, the field can instead guide intermediate
actions. Let $x_n$ be the current action hypothesis and $v_0$ the frozen
generator's velocity; here the action field is understood in the same
coordinates as $x_n$, with zeros on coordinates outside its input. A guided Euler step is
\begin{equation}
    x_{n+1}=x_n+\Delta t_n\,v_0(x_n,t_n)
       +|\Delta t_n|\,
          \mathcal{C}_b\!\left[\eta g_\theta(c_n,x_n)\right],
    \label{eq:method:euler}
\end{equation}
where $\mathcal{C}_b$ applies fixed coordinate masks and componentwise
correction bounds. The absolute time-step magnitude writes the correction
in the energy-decreasing direction even when the base sampler integrates
from noise to actions using decreasing time. The energy and its gradient
are reevaluated at each intermediate action, while a single generation
trajectory is maintained. This extends the learned preference to frozen
generative control using gradient guidance~\cite{yang2026_260702092}.

The unification lies in the trainable architecture and scalar preference
interface: function values support finite comparison, and derivatives of
the same function support continuous refinement. The action interface
determines which operation is used; each benchmark learns independent
energy parameters and uses a global gain and margin.

\FloatBarrier
\section{Experiments}
\label{sec:experiments}

Our experiments address three questions: whether a common energy architecture
can improve different frozen robot policies, how the changes are distributed
across tasks and perturbations, and which elements of action evaluation
contribute to candidate selection. The comparisons retain the same core
architecture across benchmarks while training its weights independently.

\subsection{Experimental setup}
\label{sec:experiments:setup}

\paragraph{Benchmarks and base policies.}
We consider MetaWorld MT50~\cite{pmlr-v100-yu20a}, VIMA
L1~\cite{pmlr-v202-jiang23b}, LanguageTable~\cite{lynch2022_221006407},
LIBERO Plus~\cite{Fei_2026_CVPR}, and CLIPort~\cite{pmlr-v164-shridhar22a}.
The tasks include continuous manipulation, pushing conditioned on language
instructions, and pick-and-place decisions. The base policies are the released
VIMA-200M policy, CLIPort, a language-conditioned behavior-cloning policy
for LanguageTable, and separate $\pi_{0.5}$ checkpoints for MetaWorld and
LIBERO. The LIBERO Plus adapter uses the policy pretrained for original
LIBERO. Direct denotes the corresponding frozen policy without energy-based
adaptation and serves as the local baseline for each comparison.
We omit an additional LoRA fine-tuning comparison because these base
policies are already well trained for their benchmark task families;
further low-rank policy updates would evaluate another policy-training
stage rather than the frozen-policy decision adapter studied here.

\paragraph{Coverage and metric.}
The held-out test evaluation covers 50 MT50 tasks with 50 examples each, 13 VIMA
task families with 100 each, five LanguageTable families with 50 each,
18 CLIPort variants with 100 each, and 10,030 LIBERO Plus variants.
MT50 uses the same 50 task identities in training and evaluation; its test
episodes and random seeds are held out from adapter fitting and selection.
We report task success as $100S/N$ and changes in percentage points (pp).
Each empirical success count is the number of successful episodes averaged
over five evaluation seeds and rounded to the nearest integer; the
reported rates use these rounded counts and the listed denominators.
For equal-size task groups, pooled success equals the task-macro average.
LIBERO Plus uses a variant-weighted aggregate because group sizes differ.
Its original-task, suite, and perturbation breakdowns are three views of
the same set.

CLIPort uses strict complete-task success, rather than partial cumulative
reward. The selected candidate is executed and the frozen policy continues
in closed loop to the terminal outcome. The other four benchmark protocols
evaluate adapted policies through their respective control interfaces. We fit adapters
on training data, select checkpoints and global hyperparameters on validation
data, and report Table~\ref{tab:jevvla-overall} on separate test sets.

\paragraph{Adaptation and selection.}
The core has width 128, rank 8, and 79,817 trainable parameters. Base-policy
weights and feature encoders remain frozen. Discrete interfaces use its
scalar values to rank candidates; continuous interfaces use derivatives
of the same scalar to correct actions or guide generation. Available
execution feedback, expert corrections, or demonstration preferences provide
supervision. No task-specific decision gate or teacher-policy distillation
is added. For CLIPort, we fit Full on 1,032 examples, use 258 disjoint
validation examples to select the checkpoint and global threshold, and
evaluate it on 1,800 separate test examples (18 variants $\times$ 100).
The fit and validation examples are outside the 1,800-example test set.
Full achieves 639/1800 on that test set;
all ablation differences are computed relative to this model.

\subsection{Overall performance}
\label{sec:experiments:overall}

\begin{table}[!htbp]
\centering
\small
\setlength{\tabcolsep}{5pt}
\renewcommand{\arraystretch}{1.15}
\begin{tabularx}{\linewidth}{@{}LXrJPK@{}}
\toprule
\rowcolor{jevvlaheader} \textbf{Benchmark} & \textbf{Frozen policy} & \textbf{$N$} & \textbf{Direct (\%)} & \textbf{JevVLA (\%)} & \textbf{$\Delta$ (pp)} \\
\midrule
MetaWorld MT50 & $\pi_{0.5}$--MT50 & 2,500 & 75.84 & \jevvlabest{77.12} & \jevvlagain{+1.28} \\
VIMA L1 & VIMA-200M & 1,300 & 84.77 & \jevvlabest{87.08} & \jevvlagain{+2.31} \\
LanguageTable & Language BC & 250 & 78.00 & \jevvlabest{79.60} & \jevvlagain{+1.60} \\
LIBERO Plus & $\pi_{0.5}$--LIBERO & 10,030 & 78.63 & \jevvlabest{80.15} & \jevvlagain{+1.52} \\
CLIPort & CLIPort & 1,800 & 34.22 & \jevvlabest{35.50} & \jevvlagain{+1.28} \\
\bottomrule
\end{tabularx}
\caption{Task success across five manipulation benchmarks. Each JevVLA adapter uses the corresponding frozen policy and separately trained weights. CLIPort reports strict terminal success from closed-loop execution. Evaluation protocols are specified in the text.}
\label{tab:jevvla-overall}
\end{table}

Table~\ref{tab:jevvla-overall} reports higher aggregate success for JevVLA on
all five benchmarks. MetaWorld improves from 75.84\% to
77.12\%, VIMA from 84.77\% to 87.08\%, LanguageTable from 78.00\% to
79.60\%, LIBERO Plus from 78.63\% to 80.15\%, and CLIPort from 34.22\%
to 35.50\%. The corresponding gains are 1.28, 2.31, 1.60, 1.52, and
1.28 pp. Because the base policies are frozen, these differences concern
the action decisions made with the adapter. The five comparisons play
complementary roles: they evaluate a common scoring architecture under
different observations, action parameterizations, and forms of supervision.

\begin{figure}[!htbp]
\centering
\includegraphics[width=\linewidth]{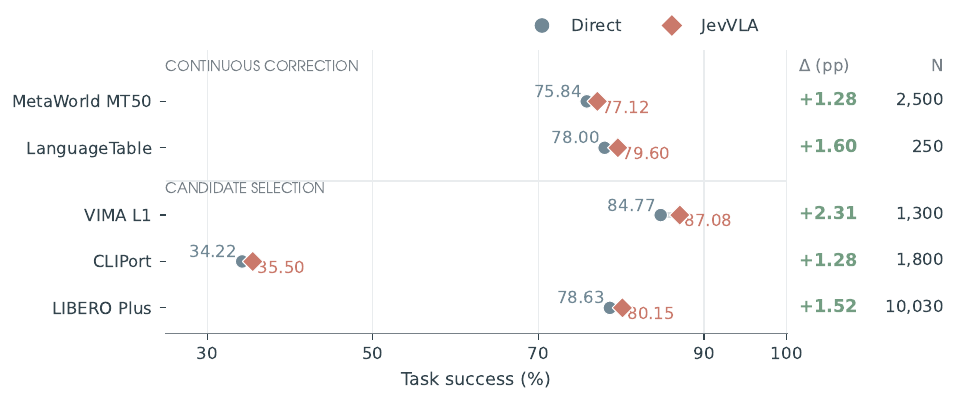}
\caption{\textbf{Task success across five manipulation benchmarks.} Circles show Direct and diamonds show JevVLA. Connectors pair results within each benchmark. The right columns give the gain in percentage points and the number of examples. Each benchmark uses independently trained adapter weights.}
\label{fig:results-overall}
\end{figure}

Figure~\ref{fig:results-overall} groups these comparisons by how the adapter
uses the learned energy: action derivatives for continuous correction,
and scalar values for candidate selection.

\paragraph{Continuous action adaptation.}
MetaWorld and LanguageTable use the energy through its action derivatives.
MT50 adds 32 successful episodes across its 50 tasks, while LanguageTable
adds four across five instruction families. The latter gain is distributed
over block-to-block, block-to-block-relative-location, and
block-to-relative-location tasks, which add one, one, and two successes,
respectively. These settings test whether a learned scalar can be useful
beyond assigning an order to a finite candidate set.

\begin{figure}[!htbp]
\centering
\includegraphics[width=\linewidth]{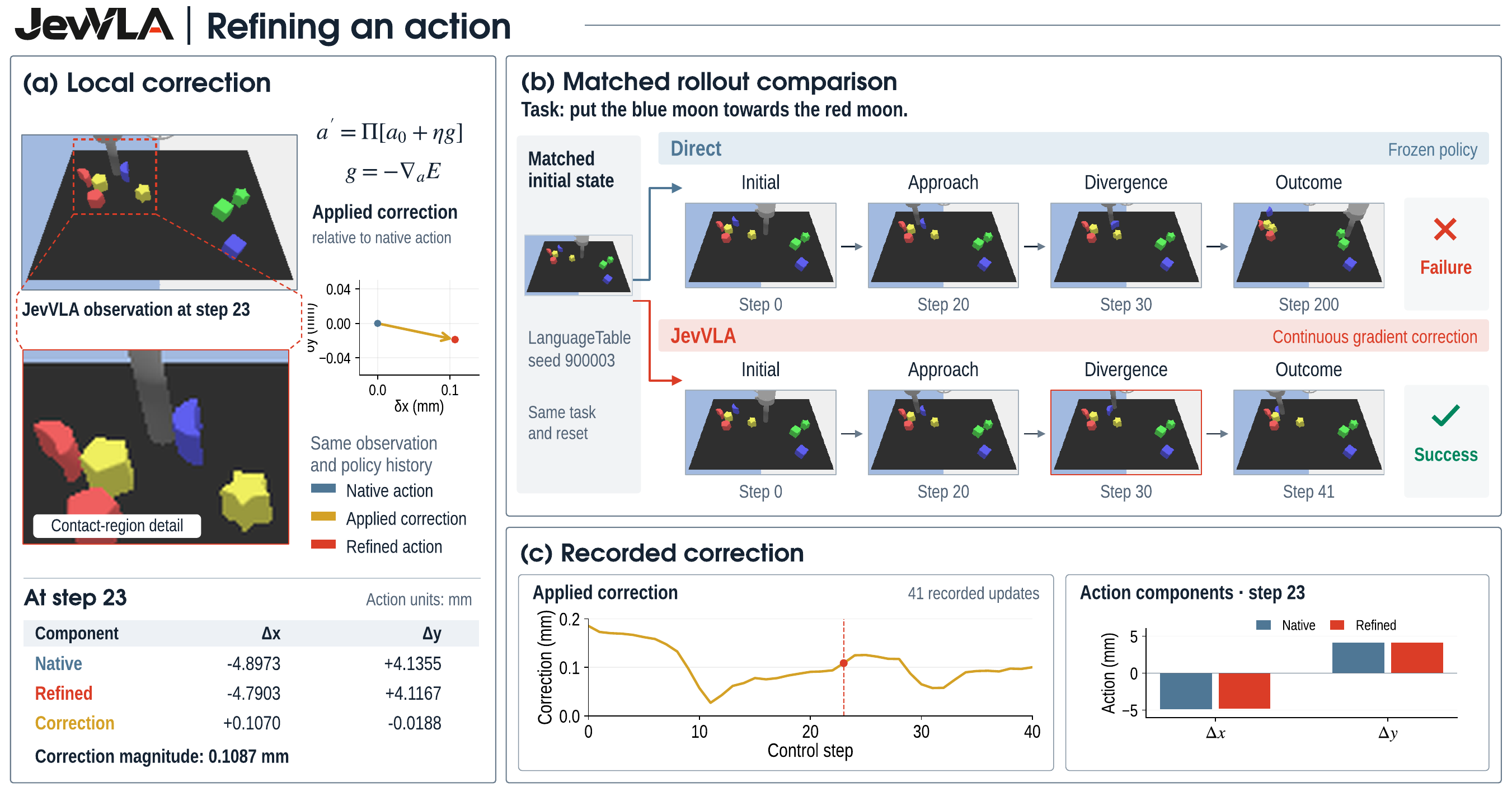}
\caption{\textbf{Continuous action refinement on LanguageTable.} Both runs start from the same recorded state (seed 900003). JevVLA succeeds at step 41; Direct reaches the 200-step limit without success. The first three rollout columns show steps 0, 20, and 30; the final column shows each actual terminal. Panel (a) compares native and refined actions on the same JevVLA observation at step 23. Panel (c) shows all 41 recorded correction magnitudes and the two action components at that step. The local vector is shown relative to the native action, in millimetres.}
\label{fig:case-continuous}
\end{figure}

Figure~\ref{fig:case-continuous} shows a separate LanguageTable rollout. Starting from the same state, JevVLA completes
the instruction in 41 steps, while Direct reaches the 200-step limit
without success. At step 23, the adapter changes the native action by
0.109\,mm. The paired trajectories and the sequence of correction
magnitudes show how small local updates change the subsequent motion.

\paragraph{Candidate-based adaptation.}
VIMA, CLIPort, and the current LIBERO Plus configuration use energy values
to compare proposals. They add 30, 23, and 152 successes, respectively.
These settings vary substantially in baseline difficulty: Direct ranges
from 34.22\% on CLIPort to 84.77\% on VIMA. The common architecture therefore
operates both when many tasks already succeed and when failures remain
frequent. Candidate construction and supervision follow each benchmark
interface. The comparisons evaluate the shared trainable architecture,
with weights learned separately for each benchmark.

The gains vary across tasks. In the task breakdowns, VIMA improves on six of 13 families, is unchanged on four,
and decreases on three. MetaWorld improves on 18 of 50 tasks, ties on 24,
and decreases on eight. CLIPort improves on 11 of 18 variants and decreases
on six; LanguageTable improves on three of five families without a decrease
in the other two. All individual task rows are retained in
Appendix~\ref{app:task-results}.

\subsection{Where adaptation helps}
\label{sec:experiments:breakdown}

\begin{figure}[!htbp]
\centering
\includegraphics[width=\linewidth]{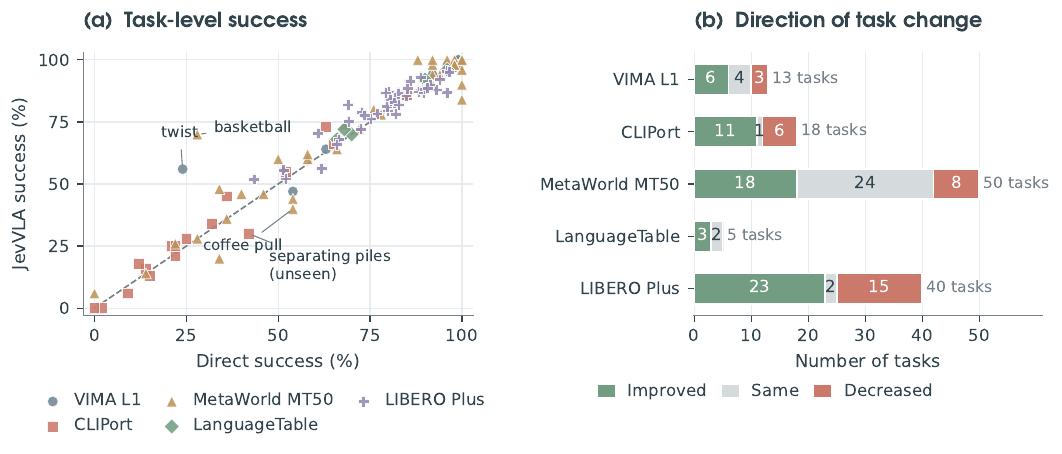}
\caption{\textbf{Task-level changes.} (a) Direct and JevVLA success for individual tasks; the dashed diagonal denotes unchanged success. (b) Numbers of improved, unchanged, and decreased tasks. LIBERO Plus groups its perturbation variants by the 40 original tasks.}
\label{fig:results-task-distribution}
\end{figure}

Figure~\ref{fig:results-task-distribution} compares individual task success
rates and counts the tasks that improve, remain unchanged, or decrease.
The distance from the diagonal shows the size of each task's change.

\paragraph{Task dependence.}
VIMA's largest increase is on \texttt{twist}, from 24\% to 56\%; this
32-example gain exceeds the net 30-example aggregate improvement because
other tasks contribute both gains and losses. In particular,
\texttt{rearrange\_then\_restore} decreases from 54\% to 47\%.
MetaWorld shows a similarly uneven pattern: basketball increases from
28\% to 70\%, while faucet opening decreases from 100\% to 84\%.
Beyond basketball, hammer increases from 34\% to 48\% and door unlocking
from 88\% to 100\%; coffee pulling and stick pulling each lose 14 pp.
A large correction on one failure-prone task can outweigh smaller
regressions elsewhere.

On CLIPort, seen-color Towers of Hanoi rises from 63\% to 73\% and
unseen-color bowl placement from 36\% to 45\%, whereas unseen-color pile
separation decreases from 42\% to 30\%. The difficult unseen-color pyramid
task remains at zero.

LanguageTable presents a smaller but more distributed change. Absolute
placement stays at 70\% and separation at 96\%; the three remaining
instruction families improve by 2, 2, and 4 pp. This differs from VIMA's
concentrated gain on \texttt{twist}.

\paragraph{Perturbation dependence.}
\begin{table}[!htbp]
\centering
\small
\setlength{\tabcolsep}{5pt}
\renewcommand{\arraystretch}{1.15}
\begin{tabularx}{\linewidth}{@{}LrJPK@{}}
\toprule
\rowcolor{jevvlaheader} \textbf{Perturbation} & \textbf{$N$} & \textbf{Direct (\%)} & \textbf{JevVLA (\%)} & \textbf{$\Delta$ (pp)} \\
\midrule
Background Textures & 1,076 & 88.94 & 86.99 & \jevvlaloss{-1.95} \\
Camera Viewpoints & 1,599 & 72.73 & 76.30 & \jevvlagain{+3.56} \\
Language Instructions & 1,537 & 79.70 & 81.33 & \jevvlagain{+1.63} \\
Light Conditions & 1,142 & 85.55 & 87.92 & \jevvlagain{+2.36} \\
Objects Layout & 1,525 & 82.03 & 84.13 & \jevvlagain{+2.10} \\
Robot Initial States & 1,550 & 65.03 & 65.55 & \jevvlagain{+0.52} \\
Sensor Noise & 1,601 & 81.57 & 83.07 & \jevvlagain{+1.50} \\
\rowcolor{jevvlatotal} All perturbations & 10,030 & 78.63 & 80.15 & \jevvlagain{+1.52} \\
\bottomrule
\end{tabularx}
\caption{LIBERO Plus success by perturbation. The seven categories partition the 10,030 variants. Task and suite summaries in the appendix regroup the same outcomes.}
\label{tab:jevvla-perturbations}
\end{table}

LIBERO Plus improves in six of seven perturbation categories
(Table~\ref{tab:jevvla-perturbations}). Camera viewpoint changes yield the
largest increase, 3.56 pp, followed by lighting changes at 2.36 pp and object
layout changes at 2.10 pp. Background texture changes instead decrease by
1.95 pp. Grouping the same examples by original suite gives gains of
2.78 pp on LIBERO-10, 3.59 pp on Goal, and 0.21 pp on Spatial, with a
0.64 pp decrease on Object.

The camera category adds 57 successes and the object-layout category adds
32, whereas background textures lose 21. These counts explain their
contributions to the aggregate, while the percentages describe the change
within each category. Robot initial-state changes add only eight successes
over 1,550 variants and remain the lowest-success group at 65.55\%.

\subsection{Ablation studies}
\label{sec:experiments:ablation}

We focus the CLIPort analysis on action selection, energy structure, and
preference supervision. All differences below are relative to Full. Configurations
that also change the training objective compare the combined effects of
the structural and training changes.

\subsubsection{Learning to select actions}
\label{sec:experiments:selection}

This comparison asks whether the available candidates benefit from a
learned contextual preference rather than a rule based only on randomness
or action geometry. Direct retains the original proposal. Uniform random
removes contextual ranking, minimum norm favors small actions, and the
nearest-alternative rule favors a small change from Direct. These controls
probe different reasons that replacing the native proposal might help.

\begin{table}[!htbp]
\centering
\small
\setlength{\tabcolsep}{5pt}
\renewcommand{\arraystretch}{1.15}
\begin{tabularx}{\linewidth}{@{}LrPK@{}}
\toprule
\rowcolor{jevvlaheader} \textbf{Selection rule} & \textbf{Count / expectation} & \textbf{Success (\%)} & \textbf{$\Delta$ vs. Direct (pp)} \\
\midrule
Direct ($c_0$) & 616/1800 & 34.22 & \jevvlaneutral{0.00} \\
Uniform random (expectation) & 616/1800 & 34.22 & \jevvlaneutral{0.00} \\
Minimum action norm & 592/1800 & 32.89 & \jevvlaloss{-1.33} \\
Nearest alternative to Direct & 600/1800 & 33.33 & \jevvlaloss{-0.89} \\
\midrule\rowcolor{jevvlaours} \textbf{JevVLA (Full)} & \textbf{639/1800} & \jevvlabest{35.50} & \jevvlagain{+1.28} \\
\bottomrule
\end{tabularx}
\caption{Selection rules on the same CLIPort candidate set. Uniform random averages the recorded candidate outcomes to obtain an analytical expectation. Differences are relative to Direct.}
\label{tab:jevvla-selection}
\end{table}

Table~\ref{tab:jevvla-selection} compares JevVLA with selectors that do not
learn contextual preferences. Uniform random selection has an expected
success rate of 34.22\%, while selecting the minimum-norm action or the
nearest alternative to Direct gives 32.89\% and 33.33\%, respectively.
JevVLA reaches 35.50\%, a gain of 1.28 pp over Direct, 2.61 pp over the
minimum-norm rule, and 2.17 pp over the nearest alternative. Learning the
relation between context and action quality gives better selections than
the size- and proximity-based rules in this comparison.

Uniform random is an analytic expectation over the candidate rollout
outcomes. Its 34.22\% matches Direct after rounding; the equality follows
from averaging these test outcomes and is not imposed by the
method. These comparisons keep candidate generation fixed and vary only
which action is selected.

\subsubsection{Structure of the energy function}
\label{sec:experiments:energy-ablation}

The energy combines conditional quadratic curvature, low-rank coupling
between action coordinates, and a nonlinear scalar residual. Their
ablations ask whether the chosen parameterization provides useful
preferences within the fixed action representation. Removing a term
changes what relationships the score can express; candidate generation
and the frozen base policy remain the same.

\begin{figure}[!htbp]
\centering
\includegraphics[width=\linewidth]{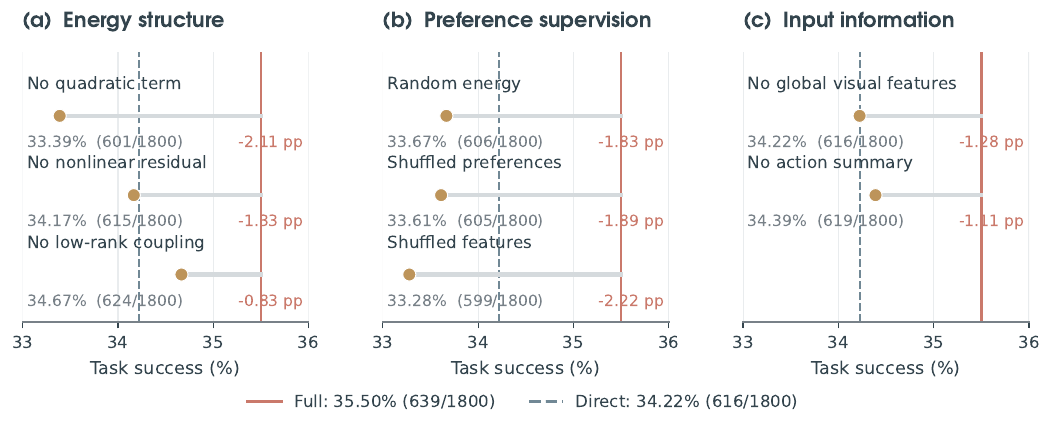}
\caption{\textbf{CLIPort ablation configurations.} Dots show success rates; differences are relative to Full (639/1800, 35.50\%, solid line). The dashed line marks Direct (616/1800, 34.22\%). The panels group energy structure, preference supervision, and scoring inputs. Full uses 1,032 fit and 258 disjoint validation examples; the 1,800 results are from a separate test set.}
\label{fig:results-ablation}
\end{figure}

Figure~\ref{fig:results-ablation} groups the structural, supervision, and
input ablations. Its reference lines mark Full and Direct, making both
the loss relative to Full and the remaining gain over Direct visible.

\begin{table}[!htbp]
\centering
\small
\setlength{\tabcolsep}{5pt}
\renewcommand{\arraystretch}{1.15}
\begin{tabularx}{\linewidth}{@{}LrPK@{}}
\toprule
\rowcolor{jevvlaheader} \textbf{Configuration} & \textbf{Success count} & \textbf{Success (\%)} & \textbf{$\Delta$ vs. Full (pp)} \\
\midrule
\rowcolor{jevvlaours} \textbf{JevVLA (Full)} & \textbf{639/1800} & \jevvlabest{35.50} & \jevvlaneutral{0.00} \\
\midrule\rowcolor{jevvlasection}\multicolumn{4}{l}{\textit{Energy structure}} \\
Without quadratic term & 601/1800 & 33.39 & \jevvlaloss{-2.11} \\
Without nonlinear residual & 615/1800 & 34.17 & \jevvlaloss{-1.33} \\
Without low-rank coupling & 624/1800 & 34.67 & \jevvlaloss{-0.83} \\
\midrule\rowcolor{jevvlasection}\multicolumn{4}{l}{\textit{Preference supervision}} \\
Randomly initialized energy & 606/1800 & 33.67 & \jevvlaloss{-1.83} \\
Shuffled preference labels & 605/1800 & 33.61 & \jevvlaloss{-1.89} \\
Shuffled input features & 599/1800 & 33.28 & \jevvlaloss{-2.22} \\
\midrule\rowcolor{jevvlasection}\multicolumn{4}{l}{\textit{Input information}} \\
Without global visual features & 616/1800 & 34.22 & \jevvlaloss{-1.28} \\
Without action summary & 619/1800 & 34.39 & \jevvlaloss{-1.11} \\
\bottomrule
\end{tabularx}
\caption{CLIPort configurations on 1,800 test examples. All gaps are relative to Full (639/1800), using 1,032 fit examples and 258 validation examples for checkpoint and threshold selection. The low-rank ablation also changes the training objective to pairwise preference learning.}
\label{tab:jevvla-ablations}
\end{table}

Removing the quadratic term gives 33.39\%, 2.11 pp below the current Full
(Table~\ref{tab:jevvla-ablations}). Removing the nonlinear residual gives
34.17\%, and removing low-rank coupling gives 34.67\%, corresponding to
gaps of 1.33 and 0.83 pp. In this comparison, the quadratic removal has
the largest deficit among the three structural variants. Removing the
residual also lowers success, which is consistent with nonlinear action
preferences carrying information beyond the structured quadratic component.
The low-rank variant probes coordinate coupling in the encoded action
space.

\subsubsection{Learning contextual preferences}
\label{sec:experiments:preference-ablation}

Preference controls examine the Jev-inspired decision principle more
directly. Their purpose is to retain an explicit scorer while disrupting
how it learns the relative quality of actions. Random initialization
removes learned preference parameters; label shuffling breaks the training
association between proposals and their preference targets; feature
shuffling breaks the corresponding association with the scoring inputs.
These controls are more specific than removing the adapter altogether.

Retaining an energy network at random initialization gives
33.67\%; shuffling preference labels gives 33.61\%, and shuffling input
features gives 33.28\%. Their gaps from Full are 1.83, 1.89, and 2.22 pp.
The differences are consistent with useful information in the learned
association between context, actions, and supervision. Random initialization is distinct
from uniform random selection: the former applies an untrained function
to the inputs and can induce a nonuniform choice distribution.

\paragraph{Scoring inputs.}
Removing global visual features yields 34.22\%, equal to Direct in aggregate
and 1.28 pp below Full. Removing the action summary yields 34.39\%, 1.11 pp
below Full. The visual comparison removes the global feature slice from the fixed
encoding, and the action-summary comparison removes the proposal summary
from the context. The remaining visual features and candidate action
encoding stay unchanged.

Removing global visual features yields 70 rescues and 70 harms, which
cancel and leave aggregate success equal to Direct. Removing local visual
features gives 35.56\%, close to Full's 35.50\%.

\begin{table}[!htbp]
\centering
\small
\setlength{\tabcolsep}{5pt}
\renewcommand{\arraystretch}{1.15}
\begin{tabularx}{\linewidth}{@{}LrrrrP@{}}
\toprule
\rowcolor{jevvlaheader} \textbf{Configuration} & \textbf{Changes} & \textbf{Rescues} & \textbf{Harms} & \textbf{Net} & \textbf{Success (\%)} \\
\midrule
Randomly initialized energy & 1,347 & 86 & 96 & -10 & 33.67 \\
Shuffled preference labels & 691 & 52 & 63 & -11 & 33.61 \\
Shuffled input features & 1,192 & 62 & 79 & -17 & 33.28 \\
\midrule\rowcolor{jevvlaours} JevVLA (Full) & 1,316 & 82 & 59 & +23 & 35.50 \\
\bottomrule
\end{tabularx}
\caption{CLIPort outcome changes relative to Direct ($N=1{,}800$). Changes counts candidate replacements. Rescues turn failures into successes; harms turn successes into failures. Net equals rescues minus harms, and total successes equal $616+\mathrm{Net}$.}
\label{tab:jevvla-interventions}
\end{table}

\paragraph{Successful corrections and disrupted successes.}
A rescue changes a Direct failure into success, while a harm changes a
Direct success into failure. For any selector, the accounting identity is
$S_{\mathrm{selector}}=S_{\mathrm{Direct}}+R-H$.
Table~\ref{tab:jevvla-interventions} shows that Full has 82 rescues and
59 harms, yielding $616+82-59=639$ successes. It changes the candidate
in 1,316 of 1,800 cases (73.11\%), but only 141 cases change their binary
outcome. The other 1,175 interventions preserve success or failure.

\begin{figure}[!htbp]
\centering
\includegraphics[width=\linewidth]{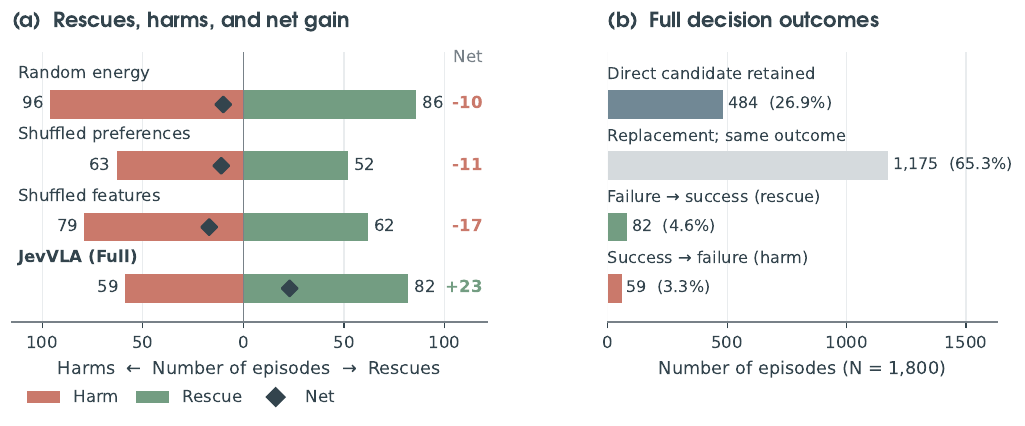}
\caption{\textbf{Rescues and harms on CLIPort.} (a) Leftward bars count Direct successes lost after replacement; rightward bars count Direct failures rescued. Diamonds indicate the net change. (b) Full retains the Direct action in 484 cases, changes the action with the same terminal outcome in 1,175, rescues 82, and harms 59. These categories sum to 1,800; rescues minus harms yield 23 additional successes.}
\label{fig:results-interventions}
\end{figure}

Figure~\ref{fig:results-interventions} separates rescues from harms for
each configuration and partitions Full's decisions into retained actions,
replacements with unchanged outcomes, rescues, and harms.

The randomly initialized energy has slightly more rescues, 86, but also
96 harms. Relative to those aggregate counts, Full gives up four rescues
while reducing harms by 37, for a net improvement of 33 successes. The
shuffled-label configuration has 52 rescues and 63 harms; Full gains
30 rescues and incurs four fewer harms, explaining its 34-example
advantage.

\begin{figure}[!htbp]
\centering
\includegraphics[width=\linewidth]{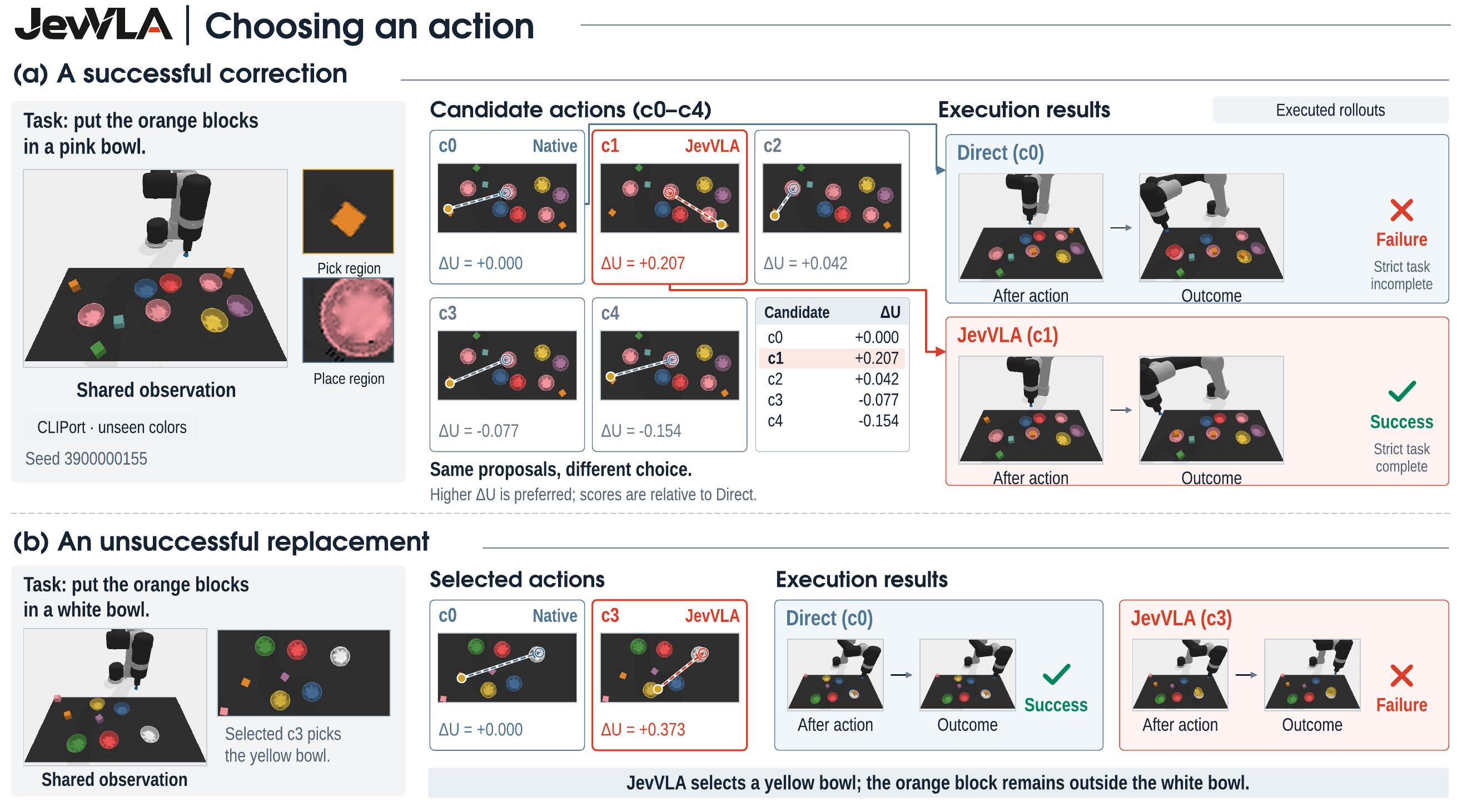}
\caption{\textbf{Candidate selection on CLIPort.} The upper case (seed 3900000155) changes the first proposal from $c_0$ to $c_1$; frozen-policy continuation then places both orange blocks in pink bowls. The lower case (seed 3900000148) selects $c_3$, which moves a yellow bowl into the white bowl while leaving the orange block outside. Candidate overlays show recorded pick and place positions on the same initial heightmap. Displayed scores are $\Delta U=U(c_i)-U(c_0)$, with $E=-U$. Camera frames replay the four recorded branches with the frozen base policy. The aggregate CLIPort evaluation covers 1,800 test examples.}
\label{fig:case-candidate}
\end{figure}

Figure~\ref{fig:case-candidate} shows one rescue and one harm from recorded
CLIPort branches. In the rescue, both initial proposals move an
orange block toward a pink bowl, but their subsequent continuations lead
to different terminal outcomes. In the harm, the selected proposal moves
a yellow bowl into the white bowl and leaves the orange block outside.
The candidate scores and execution frames connect each decision to the
resulting scene.

JevVLA draws on Jev's explicit representation of alternatives and their
relative preferences. These experiments examine how learning those
preferences from supervision affects action selection.

\subsection{Comparison methods}
\label{sec:experiments:competitors}

We compare JevVLA with other methods for evaluating or selecting actions
from frozen policies. On MetaWorld, we implement a shared linear value
probe and the RAS decoding rule with the same frozen $\pi_{0.5}$ policy.
On CLIPort, logistic regression, a shared MLP, TabM, FT-Transformer, and
TabNet serve as terminal-outcome classifiers for branch selection.

\begin{table}[!htbp]
\centering\small
\setlength{\tabcolsep}{6pt}\renewcommand{\arraystretch}{1.14}
\begin{tabularx}{\linewidth}{@{}Lrr@{}}
\toprule
\rowcolor{jevvlaheader} \textbf{Method} & \textbf{Successes} & \textbf{Success (\%)} \\
\midrule
\rowcolor{jevvladirect} Direct ($\pi_{0.5}$) & 1896/2500 & 75.84 \\
Shared linear value probe~\cite{zhang2026frozenvalueprobe} & 1839/2500 & 73.56 \\
RSS RAS decoding rule~\cite{zhan2026stablelanguage} & 1168/2500 & 46.72 \\
\rowcolor{jevvlaours} \textbf{JevVLA (Full)} & \textbf{1928/2500} & \jevvlabest{77.12} \\
\bottomrule
\end{tabularx}
\caption{Method comparisons on MetaWorld MT50 (50 tasks, 50 test episodes each). All methods use the frozen $\pi_{0.5}$ policy.}
\label{tab:jevvla-meta-competitors}
\end{table}

\begin{table}[!htbp]
\centering\small
\setlength{\tabcolsep}{6pt}\renewcommand{\arraystretch}{1.14}
\begin{tabularx}{\linewidth}{@{}Lrr@{}}
\toprule
\rowcolor{jevvlaheader} \textbf{Method} & \textbf{Successes} & \textbf{Success (\%)} \\
\midrule
\rowcolor{jevvladirect} Direct (CLIPort) & 616/1800 & 34.22 \\
UA-CLIPort~\cite{wu2024uncertaintyaware} & 607/1800 & 33.72 \\
Logistic regression & 623/1800 & 34.61 \\
Shared MLP & 617/1800 & 34.28 \\
TabM~\cite{gorishniy2025tabm} & 627/1800 & 34.83 \\
FT-Transformer~\cite{gorishniy2021revisiting} & 624/1800 & 34.67 \\
TabNet~\cite{arik2021tabnet} & 614/1800 & 34.11 \\
\rowcolor{jevvlaours} \textbf{JevVLA (Full)} & \textbf{639/1800} & \jevvlabest{35.50} \\
\bottomrule
\end{tabularx}
\caption{Method comparisons on the 18 CLIPort variants (100 test episodes each). UA-CLIPort uses an online 11$\times$11 grasp rule; no matched online Direct run is available for this configuration.}
\label{tab:jevvla-cliport-competitors}
\end{table}

\FloatBarrier

On MetaWorld, JevVLA succeeds in 1,928 of 2,500 episodes, ahead of Direct
(1,896), the linear probe (1,839), and the RAS rule (1,168). The probe
still leads on basketball (40/50 versus 35/50) and coffee pulling
(37/50 versus 20/50).

On CLIPort, TabM is the strongest alternative in this set at 627/1,800.
JevVLA obtains 639/1,800, 12 more completed tasks than TabM. Logistic
regression and FT-Transformer obtain 623 and 624 successes, while the
shared MLP and TabNet obtain 617 and 614. The online UA-CLIPort rule
obtains 607 successes on the same 18-variant, 100-per-variant task mix.
Because its evaluation has no matched online Direct run, we report its
count without interpreting the difference from the Direct row as a
paired effect. Appendix~\ref{app:alternative-methods} lists all
per-task counts for these comparisons.

\FloatBarrier

\section{Conclusion}
\label{sec:conclusion}

We presented JevVLA, a lightweight adaptation framework inspired by Jev's
explicit comparison of alternatives and their relative preferences. Its
central contribution is one trainable action preference core for two VLA
interfaces. With the base policy frozen, scalar energy values rank discrete
proposals, while gradients of the same energy refine continuous actions or
guide their generation. The core has 79,817 trainable parameters, with
weights learned separately for each benchmark.

Across five manipulation benchmarks, JevVLA improves aggregate task success
over the corresponding frozen policies. The CLIPort ablations show the
contributions of the energy structure, preference supervision, and scoring
inputs. The execution examples illustrate both how small corrections
change a rollout and how candidate ranking changes the final outcome.
Together, these results support a shared action preference architecture
for improving both candidate selection and continuous control without
updating the base policy.

\bibliographystyle{unsrtnat}
\bibliography{references}
\clearpage
\appendix
\section{Reproducibility details}
\label{app:reproducibility}

The trainable core is the width-128, rank-8 scalar energy in
Section~\ref{sec:method:core}; its 79,817 parameters are initialized and
trained separately for each benchmark. The frozen policies, feature
encoders, and benchmark-specific input/output codecs are outside this
parameter count. MetaWorld supervision combines physically executed branch
returns with scripted expert corrections. VIMA uses executed candidate
outcomes and expert alignment. LanguageTable uses its scripted oracle at
observed and later states. LIBERO Plus uses future-action proximity to
official demonstration trajectories. CLIPort uses terminal outcomes of
physically executed candidates. These sources yield preference targets
or local gradient targets as defined in Section~\ref{sec:method:training}.

For the selected CLIPort comparison, the adapter is fit on 1,032 examples;
258 other examples select the checkpoint and global Direct margin, and
1,800 separate closed-loop test episodes supply the reported strict-success
counts. The test set contains 100 examples for each of 18 task variants.
The other reported test-set sizes are 2,500 for MetaWorld, 1,300 for VIMA,
250 for LanguageTable, and 10,030 for LIBERO Plus, as detailed in
Section~\ref{sec:experiments:setup}. Candidate sets and rollout horizons
follow the corresponding benchmark interfaces.

\paragraph{Optimization.}
All adapters use AdamW with weight decay $10^{-4}$ and a gradient-norm cap
of 1. The table lists the energy-adapter settings; it does not include
optimization of the frozen base policies. Epochs in parentheses identify
the checkpoint used after validation selection.
\begin{table}[!htbp]
\centering\footnotesize
\setlength{\tabcolsep}{4pt}
\renewcommand{\arraystretch}{1.13}
\begin{tabularx}{\linewidth}{@{}p{0.17\linewidth}p{0.43\linewidth}X@{}}
\toprule
\rowcolor{jevvlaheader}\textbf{Benchmark} & \textbf{Adapter optimization} & \textbf{Target and inference settings} \\
\midrule
MetaWorld MT50 & Learning rate $2\times10^{-4}$, batch 128, 8 initial epochs; $3\times10^{-5}$, batch 128, 3 continuation epochs (1). & Expert shift $\alpha=0.25$; $\sigma(t)=0.2+0.8t$; correction gain $\eta=0.08$; component clip $0.12$. \\
VIMA L1 & Learning rate $3\times10^{-4}$, batch 64, 80 epochs (80); seed 5137. & Five training candidates; target temperature $\tau_y=0.235$; native-prior scale $\tau_0=20$; energy gain $\beta=4$. \\
LanguageTable & Learning rate $3\times10^{-4}$; each step uses 16 initial anchors and up to 128 future states; 80-epoch selection, 6-epoch refit; seed 260930. & Initial CE and DSM weights $1:1$, future-state DSM weight $0.2$; $\sigma(t)=0.2+0.8t$; $\eta=0.0004$. \\
LIBERO Plus & Learning rate $3\times10^{-4}$, batch 128, up to 80 epochs (46); seed 5137. & Two candidates; $\tau_y=0.0312$; $\beta=0.25$; global Direct margin $m=0.005$. \\
CLIPort & Learning rate $3\times10^{-4}$, batch 128, up to 80 epochs for CE Full; pairwise ablations use up to 50 epochs. & Five executed candidates; $\beta=1$; the checkpoint and global Direct margin $m$ are selected on the 258-example validation set. \\
\bottomrule
\end{tabularx}
\caption{Energy-adapter hyperparameters. A parenthesized epoch is the selected checkpoint; other epoch counts are training limits.}
\label{tab:repro-hyperparameters}
\end{table}

The temperature $\tau_y$ controls a demonstration- or expert-derived
target distribution, whereas $\tau_0$ rescales a native proposal score.
CLIPort and LIBERO Plus do not use a native proposal prior, so $\tau_0$
does not enter their selection logits. The mixture masses $\rho_k$ in
Eq.~\eqref{eq:method:mixture} are normalized per anchor from measured
branch return, terminal success, and future expert alignment; they are
not one global learned coefficient. Discrete selection does not use
$\eta$, $\sigma$, or $\alpha$; continuous correction does not use the
candidate margin $m$.

\paragraph{Compute.}
The experiments used a single server with eight NVIDIA A100-SXM4 GPUs
(80\,GB each) and two Intel Xeon Platinum 8338C CPUs. The small VIMA,
LanguageTable, LIBERO Plus, and CLIPort energy adapters fit on CPU; the
MetaWorld energy continuation uses one A100. Frozen-policy inference and
simulator rollouts use GPU services, while no adapter requires multi-GPU
optimization.

\section{Task-level results}
\label{app:task-results}
These tables report the full task breakdowns. Task names are shortened only by replacing separators with spaces.

\subsection{VIMA L1}
\begingroup\small
\setlength{\tabcolsep}{4pt}\renewcommand{\arraystretch}{1.12}
\begin{longtable}{@{}p{0.49\linewidth}rJPK@{}}
\caption{VIMA L1: task counts and success rates.}\\
\toprule\rowcolor{jevvlaheader} \textbf{Task} & \textbf{$N$} & \textbf{Direct (\%)} & \textbf{JevVLA (\%)} & \textbf{$\Delta$ (pp)} \\
\midrule\endfirsthead
\toprule\rowcolor{jevvlaheader} \textbf{Task (continued)} & \textbf{$N$} & \textbf{Direct (\%)} & \textbf{JevVLA (\%)} & \textbf{$\Delta$ (pp)} \\
\midrule\endhead
\bottomrule\endfoot
follow order & 100 & 85.00 & 85.00 & \jevvlaneutral{0.00} \\
manipulate old neighbor & 100 & 92.00 & 94.00 & \jevvlagain{+2.00} \\
novel adj & 100 & 96.00 & 98.00 & \jevvlagain{+2.00} \\
novel noun & 100 & 98.00 & 97.00 & \jevvlaloss{-1.00} \\
pick in order then restore & 100 & 63.00 & 64.00 & \jevvlagain{+1.00} \\
rearrange & 100 & 99.00 & 100.00 & \jevvlagain{+1.00} \\
rearrange then restore & 100 & 54.00 & 47.00 & \jevvlaloss{-7.00} \\
rotate & 100 & 98.00 & 98.00 & \jevvlaneutral{0.00} \\
same shape & 100 & 98.00 & 98.00 & \jevvlaneutral{0.00} \\
scene understanding & 100 & 98.00 & 98.00 & \jevvlaneutral{0.00} \\
sweep without exceeding & 100 & 98.00 & 97.00 & \jevvlaloss{-1.00} \\
twist & 100 & 24.00 & 56.00 & \jevvlagain{+32.00} \\
visual manipulation & 100 & 99.00 & 100.00 & \jevvlagain{+1.00} \\
\textbf{Total} & 1,300 & 84.77 & 87.08 & \jevvlagain{+2.31} \\
\end{longtable}\endgroup

\subsection{CLIPort}
\begingroup\small
\setlength{\tabcolsep}{4pt}\renewcommand{\arraystretch}{1.12}
\begin{longtable}{@{}p{0.49\linewidth}rJPK@{}}
\caption{CLIPort: task counts and success rates.}\\
\toprule\rowcolor{jevvlaheader} \textbf{Task} & \textbf{$N$} & \textbf{Direct (\%)} & \textbf{JevVLA (\%)} & \textbf{$\Delta$ (pp)} \\
\midrule\endfirsthead
\toprule\rowcolor{jevvlaheader} \textbf{Task (continued)} & \textbf{$N$} & \textbf{Direct (\%)} & \textbf{JevVLA (\%)} & \textbf{$\Delta$ (pp)} \\
\midrule\endhead
\bottomrule\endfoot
align rope & 100 & 9.00 & 6.00 & \jevvlaloss{-3.00} \\
assembling kits seq seen colors & 100 & 22.00 & 21.00 & \jevvlaloss{-1.00} \\
assembling kits seq unseen colors & 100 & 2.00 & 0.00 & \jevvlaloss{-2.00} \\
packing boxes pairs seen colors & 100 & 52.00 & 55.00 & \jevvlagain{+3.00} \\
packing boxes pairs unseen colors & 100 & 14.00 & 16.00 & \jevvlagain{+2.00} \\
packing seen google objects group & 100 & 21.00 & 25.00 & \jevvlagain{+4.00} \\
packing seen google objects seq & 100 & 12.00 & 18.00 & \jevvlagain{+6.00} \\
packing shapes & 100 & 32.00 & 34.00 & \jevvlagain{+2.00} \\
packing unseen google objects group & 100 & 25.00 & 28.00 & \jevvlagain{+3.00} \\
packing unseen google objects seq & 100 & 15.00 & 13.00 & \jevvlaloss{-2.00} \\
put block in bowl seen colors & 100 & 99.00 & 98.00 & \jevvlaloss{-1.00} \\
put block in bowl unseen colors & 100 & 36.00 & 45.00 & \jevvlagain{+9.00} \\
separating piles seen colors & 100 & 65.00 & 66.00 & \jevvlagain{+1.00} \\
separating piles unseen colors & 100 & 42.00 & 30.00 & \jevvlaloss{-12.00} \\
stack block pyramid seq seen colors & 100 & 85.00 & 86.00 & \jevvlagain{+1.00} \\
stack block pyramid seq unseen colors & 100 & 0.00 & 0.00 & \jevvlaneutral{0.00} \\
towers of hanoi seq seen colors & 100 & 63.00 & 73.00 & \jevvlagain{+10.00} \\
towers of hanoi seq unseen colors & 100 & 22.00 & 25.00 & \jevvlagain{+3.00} \\
\textbf{Total} & 1,800 & 34.22 & 35.50 & \jevvlagain{+1.28} \\
\end{longtable}\endgroup

\subsection{MetaWorld MT50}
\begingroup\small
\setlength{\tabcolsep}{4pt}\renewcommand{\arraystretch}{1.12}
\begin{longtable}{@{}p{0.49\linewidth}rJPK@{}}
\caption{MetaWorld MT50: task counts and success rates.}\\
\toprule\rowcolor{jevvlaheader} \textbf{Task} & \textbf{$N$} & \textbf{Direct (\%)} & \textbf{JevVLA (\%)} & \textbf{$\Delta$ (pp)} \\
\midrule\endfirsthead
\toprule\rowcolor{jevvlaheader} \textbf{Task (continued)} & \textbf{$N$} & \textbf{Direct (\%)} & \textbf{JevVLA (\%)} & \textbf{$\Delta$ (pp)} \\
\midrule\endhead
\bottomrule\endfoot
assembly & 50 & 98.00 & 98.00 & \jevvlaneutral{0.00} \\
basketball & 50 & 28.00 & 70.00 & \jevvlagain{+42.00} \\
bin picking & 50 & 54.00 & 44.00 & \jevvlaloss{-10.00} \\
box close & 50 & 58.00 & 60.00 & \jevvlagain{+2.00} \\
button press topdown & 50 & 98.00 & 98.00 & \jevvlaneutral{0.00} \\
button press topdown wall & 50 & 100.00 & 100.00 & \jevvlaneutral{0.00} \\
button press & 50 & 100.00 & 100.00 & \jevvlaneutral{0.00} \\
button press wall & 50 & 100.00 & 100.00 & \jevvlaneutral{0.00} \\
coffee button & 50 & 100.00 & 100.00 & \jevvlaneutral{0.00} \\
coffee pull & 50 & 54.00 & 40.00 & \jevvlaloss{-14.00} \\
coffee push & 50 & 100.00 & 100.00 & \jevvlaneutral{0.00} \\
dial turn & 50 & 36.00 & 36.00 & \jevvlaneutral{0.00} \\
disassemble & 50 & 78.00 & 78.00 & \jevvlaneutral{0.00} \\
door close & 50 & 92.00 & 94.00 & \jevvlagain{+2.00} \\
door lock & 50 & 90.00 & 90.00 & \jevvlaneutral{0.00} \\
door open & 50 & 100.00 & 90.00 & \jevvlaloss{-10.00} \\
door unlock & 50 & 88.00 & 100.00 & \jevvlagain{+12.00} \\
drawer close & 50 & 100.00 & 100.00 & \jevvlaneutral{0.00} \\
drawer open & 50 & 100.00 & 100.00 & \jevvlaneutral{0.00} \\
faucet close & 50 & 92.00 & 98.00 & \jevvlagain{+6.00} \\
faucet open & 50 & 100.00 & 84.00 & \jevvlaloss{-16.00} \\
hammer & 50 & 34.00 & 48.00 & \jevvlagain{+14.00} \\
hand insert & 50 & 100.00 & 100.00 & \jevvlaneutral{0.00} \\
handle press side & 50 & 82.00 & 84.00 & \jevvlagain{+2.00} \\
handle press & 50 & 90.00 & 90.00 & \jevvlaneutral{0.00} \\
handle pull side & 50 & 0.00 & 6.00 & \jevvlagain{+6.00} \\
handle pull & 50 & 14.00 & 14.00 & \jevvlaneutral{0.00} \\
lever pull & 50 & 46.00 & 46.00 & \jevvlaneutral{0.00} \\
peg insert side & 50 & 92.00 & 100.00 & \jevvlagain{+8.00} \\
peg unplug side & 50 & 92.00 & 90.00 & \jevvlaloss{-2.00} \\
pick out of hole & 50 & 22.00 & 26.00 & \jevvlagain{+4.00} \\
pick place & 50 & 66.00 & 64.00 & \jevvlaloss{-2.00} \\
pick place wall & 50 & 96.00 & 98.00 & \jevvlagain{+2.00} \\
plate slide back side & 50 & 100.00 & 100.00 & \jevvlaneutral{0.00} \\
plate slide back & 50 & 96.00 & 100.00 & \jevvlagain{+4.00} \\
plate slide side & 50 & 100.00 & 96.00 & \jevvlaloss{-4.00} \\
plate slide & 50 & 98.00 & 100.00 & \jevvlagain{+2.00} \\
push back & 50 & 40.00 & 46.00 & \jevvlagain{+6.00} \\
push & 50 & 100.00 & 100.00 & \jevvlaneutral{0.00} \\
push wall & 50 & 82.00 & 82.00 & \jevvlaneutral{0.00} \\
reach & 50 & 28.00 & 28.00 & \jevvlaneutral{0.00} \\
reach wall & 50 & 72.00 & 72.00 & \jevvlaneutral{0.00} \\
shelf place & 50 & 58.00 & 62.00 & \jevvlagain{+4.00} \\
soccer & 50 & 68.00 & 74.00 & \jevvlagain{+6.00} \\
stick pull & 50 & 34.00 & 20.00 & \jevvlaloss{-14.00} \\
stick push & 50 & 50.00 & 60.00 & \jevvlagain{+10.00} \\
sweep into & 50 & 94.00 & 94.00 & \jevvlaneutral{0.00} \\
sweep & 50 & 76.00 & 80.00 & \jevvlagain{+4.00} \\
window close & 50 & 98.00 & 98.00 & \jevvlaneutral{0.00} \\
window open & 50 & 98.00 & 98.00 & \jevvlaneutral{0.00} \\
\textbf{Total} & 2,500 & 75.84 & 77.12 & \jevvlagain{+1.28} \\
\end{longtable}\endgroup

\subsection{LanguageTable}
\begingroup\small
\setlength{\tabcolsep}{4pt}\renewcommand{\arraystretch}{1.12}
\begin{longtable}{@{}p{0.49\linewidth}rJPK@{}}
\caption{LanguageTable: task counts and success rates.}\\
\toprule\rowcolor{jevvlaheader} \textbf{Task} & \textbf{$N$} & \textbf{Direct (\%)} & \textbf{JevVLA (\%)} & \textbf{$\Delta$ (pp)} \\
\midrule\endfirsthead
\toprule\rowcolor{jevvlaheader} \textbf{Task (continued)} & \textbf{$N$} & \textbf{Direct (\%)} & \textbf{JevVLA (\%)} & \textbf{$\Delta$ (pp)} \\
\midrule\endhead
\bottomrule\endfoot
blocktoabsolutelocation & 50 & 70.00 & 70.00 & \jevvlaneutral{0.00} \\
blocktoblock & 50 & 90.00 & 92.00 & \jevvlagain{+2.00} \\
blocktoblockrelativelocation & 50 & 66.00 & 68.00 & \jevvlagain{+2.00} \\
blocktorelativelocation & 50 & 68.00 & 72.00 & \jevvlagain{+4.00} \\
separate & 50 & 96.00 & 96.00 & \jevvlaneutral{0.00} \\
\textbf{Total} & 250 & 78.00 & 79.60 & \jevvlagain{+1.60} \\
\end{longtable}\endgroup

\subsection{LIBERO Plus: original tasks}
\begingroup\small
\setlength{\tabcolsep}{4pt}\renewcommand{\arraystretch}{1.12}
\begin{longtable}{@{}p{0.49\linewidth}rJPK@{}}
\caption{LIBERO Plus: original tasks: task counts and success rates.}\\
\toprule\rowcolor{jevvlaheader} \textbf{Task} & \textbf{$N$} & \textbf{Direct (\%)} & \textbf{JevVLA (\%)} & \textbf{$\Delta$ (pp)} \\
\midrule\endfirsthead
\toprule\rowcolor{jevvlaheader} \textbf{Task (continued)} & \textbf{$N$} & \textbf{Direct (\%)} & \textbf{JevVLA (\%)} & \textbf{$\Delta$ (pp)} \\
\midrule\endhead
\bottomrule\endfoot
libero 10: KITCHEN SCENE3 turn on the stove and put the moka pot on it & 281 & 69.04 & 81.85 & \jevvlagain{+12.81} \\
libero 10: KITCHEN SCENE4 put the black bowl in the bottom drawer of the cabinet and close it & 304 & 76.97 & 78.29 & \jevvlagain{+1.32} \\
libero 10: KITCHEN SCENE6 put the yellow and white mug in the microwave and close it & 278 & 66.55 & 67.99 & \jevvlagain{+1.44} \\
libero 10: KITCHEN SCENE8 put both moka pots on the stove & 233 & 61.80 & 56.22 & \jevvlaloss{-5.58} \\
libero 10: LIVING ROOM SCENE1 put both the alphabet soup and the cream cheese box in the basket & 194 & 79.90 & 79.38 & \jevvlaloss{-0.52} \\
libero 10: LIVING ROOM SCENE2 put both the alphabet soup and the tomato sauce in the basket & 212 & 82.08 & 83.02 & \jevvlagain{+0.94} \\
libero 10: LIVING ROOM SCENE2 put both the cream cheese box and the butter in the basket & 210 & 75.24 & 76.19 & \jevvlagain{+0.95} \\
libero 10: LIVING ROOM SCENE5 put the white mug on the left plate and put the yellow and white mug on the right plate & 287 & 72.47 & 72.13 & \jevvlaloss{-0.35} \\
libero 10: LIVING ROOM SCENE6 put the white mug on the plate and put the chocolate pudding to the right of the plate & 283 & 72.79 & 78.80 & \jevvlagain{+6.01} \\
libero 10: STUDY SCENE1 pick up the book and place it in the back compartment of the caddy & 237 & 43.46 & 51.90 & \jevvlagain{+8.44} \\
libero goal: open the middle drawer of the cabinet & 268 & 73.51 & 77.61 & \jevvlagain{+4.10} \\
libero goal: open the top drawer and put the bowl inside & 324 & 79.63 & 81.17 & \jevvlagain{+1.54} \\
libero goal: push the plate to the front of the stove & 270 & 51.48 & 55.56 & \jevvlagain{+4.07} \\
libero goal: put the bowl on the plate & 235 & 81.70 & 82.13 & \jevvlagain{+0.43} \\
libero goal: put the bowl on the stove & 156 & 82.69 & 86.54 & \jevvlagain{+3.85} \\
libero goal: put the bowl on top of the cabinet & 243 & 60.91 & 70.37 & \jevvlagain{+9.47} \\
libero goal: put the cream cheese in the bowl & 291 & 84.88 & 85.57 & \jevvlagain{+0.69} \\
libero goal: put the wine bottle on the rack & 302 & 69.21 & 75.17 & \jevvlagain{+5.96} \\
libero goal: put the wine bottle on top of the cabinet & 270 & 65.93 & 65.93 & \jevvlaneutral{0.00} \\
libero goal: turn on the stove & 232 & 80.17 & 87.07 & \jevvlagain{+6.90} \\
libero object: pick up the alphabet soup and place it in the basket & 242 & 95.87 & 95.45 & \jevvlaloss{-0.41} \\
libero object: pick up the bbq sauce and place it in the basket & 292 & 85.96 & 91.44 & \jevvlagain{+5.48} \\
libero object: pick up the butter and place it in the basket & 273 & 95.97 & 86.81 & \jevvlaloss{-9.16} \\
libero object: pick up the chocolate pudding and place it in the basket & 257 & 88.33 & 87.16 & \jevvlaloss{-1.17} \\
libero object: pick up the cream cheese and place it in the basket & 173 & 89.60 & 86.71 & \jevvlaloss{-2.89} \\
libero object: pick up the ketchup and place it in the basket & 214 & 81.78 & 84.58 & \jevvlagain{+2.80} \\
libero object: pick up the milk and place it in the basket & 272 & 80.51 & 84.19 & \jevvlagain{+3.68} \\
libero object: pick up the orange juice and place it in the basket & 280 & 90.00 & 88.21 & \jevvlaloss{-1.79} \\
libero object: pick up the salad dressing and place it in the basket & 264 & 96.59 & 95.08 & \jevvlaloss{-1.52} \\
libero object: pick up the tomato sauce and place it in the basket & 251 & 94.02 & 92.03 & \jevvlaloss{-1.99} \\
libero spatial: pick up the black bowl between the plate and the ramekin and place it on the plate & 231 & 93.07 & 87.88 & \jevvlaloss{-5.19} \\
libero spatial: pick up the black bowl from table center and place it on the plate & 267 & 91.39 & 89.89 & \jevvlaloss{-1.50} \\
libero spatial: pick up the black bowl in the top drawer of the wooden cabinet and place it on the plate & 280 & 52.14 & 52.14 & \jevvlaneutral{0.00} \\
libero spatial: pick up the black bowl next to the cookie box and place it on the plate & 232 & 79.31 & 86.64 & \jevvlagain{+7.33} \\
libero spatial: pick up the black bowl next to the plate and place it on the plate & 180 & 81.67 & 82.78 & \jevvlagain{+1.11} \\
libero spatial: pick up the black bowl next to the ramekin and place it on the plate & 198 & 82.83 & 81.82 & \jevvlaloss{-1.01} \\
libero spatial: pick up the black bowl on the cookie box and place it on the plate & 246 & 90.65 & 88.62 & \jevvlaloss{-2.03} \\
libero spatial: pick up the black bowl on the ramekin and place it on the plate & 256 & 82.03 & 78.12 & \jevvlaloss{-3.91} \\
libero spatial: pick up the black bowl on the stove and place it on the plate & 269 & 88.85 & 92.94 & \jevvlagain{+4.09} \\
libero spatial: pick up the black bowl on the wooden cabinet and place it on the plate & 243 & 85.19 & 88.48 & \jevvlagain{+3.29} \\
\textbf{Total} & 10,030 & 78.63 & 80.15 & \jevvlagain{+1.52} \\
\end{longtable}\endgroup

\subsection{LIBERO Plus: perturbations}
\begingroup\small
\setlength{\tabcolsep}{4pt}\renewcommand{\arraystretch}{1.12}
\begin{longtable}{@{}p{0.49\linewidth}rJPK@{}}
\caption{LIBERO Plus: perturbations: task counts and success rates.}\\
\toprule\rowcolor{jevvlaheader} \textbf{Task} & \textbf{$N$} & \textbf{Direct (\%)} & \textbf{JevVLA (\%)} & \textbf{$\Delta$ (pp)} \\
\midrule\endfirsthead
\toprule\rowcolor{jevvlaheader} \textbf{Task (continued)} & \textbf{$N$} & \textbf{Direct (\%)} & \textbf{JevVLA (\%)} & \textbf{$\Delta$ (pp)} \\
\midrule\endhead
\bottomrule\endfoot
Background Textures & 1,076 & 88.94 & 86.99 & \jevvlaloss{-1.95} \\
Camera Viewpoints & 1,599 & 72.73 & 76.30 & \jevvlagain{+3.56} \\
Language Instructions & 1,537 & 79.70 & 81.33 & \jevvlagain{+1.63} \\
Light Conditions & 1,142 & 85.55 & 87.92 & \jevvlagain{+2.36} \\
Objects Layout & 1,525 & 82.03 & 84.13 & \jevvlagain{+2.10} \\
Robot Initial States & 1,550 & 65.03 & 65.55 & \jevvlagain{+0.52} \\
Sensor Noise & 1,601 & 81.57 & 83.07 & \jevvlagain{+1.50} \\
\textbf{Total} & 10,030 & 78.63 & 80.15 & \jevvlagain{+1.52} \\
\end{longtable}\endgroup

\subsection{LIBERO Plus: suites}
\begingroup\small
\setlength{\tabcolsep}{4pt}\renewcommand{\arraystretch}{1.12}
\begin{longtable}{@{}p{0.49\linewidth}rJPK@{}}
\caption{LIBERO Plus: suites: task counts and success rates.}\\
\toprule\rowcolor{jevvlaheader} \textbf{Task} & \textbf{$N$} & \textbf{Direct (\%)} & \textbf{JevVLA (\%)} & \textbf{$\Delta$ (pp)} \\
\midrule\endfirsthead
\toprule\rowcolor{jevvlaheader} \textbf{Task (continued)} & \textbf{$N$} & \textbf{Direct (\%)} & \textbf{JevVLA (\%)} & \textbf{$\Delta$ (pp)} \\
\midrule\endhead
\bottomrule\endfoot
libero 10 & 2,519 & 69.91 & 72.69 & \jevvlagain{+2.78} \\
libero goal & 2,591 & 72.67 & 76.26 & \jevvlagain{+3.59} \\
libero object & 2,518 & 89.91 & 89.28 & \jevvlaloss{-0.64} \\
libero spatial & 2,402 & 82.39 & 82.60 & \jevvlagain{+0.21} \\
\textbf{Total} & 10,030 & 78.63 & 80.15 & \jevvlagain{+1.52} \\
\end{longtable}\endgroup

\section{Comparison method results by task}
\label{app:alternative-methods}
The following tables give the per-task counts underlying Tables~\ref{tab:jevvla-meta-competitors} and~\ref{tab:jevvla-cliport-competitors}.

\begingroup\footnotesize
\setlength{\tabcolsep}{4pt}\renewcommand{\arraystretch}{1.10}
\begin{longtable}{@{}p{0.48\linewidth}rrrr@{}}
\caption{MetaWorld MT50: comparison methods by task.}\label{tab:jevvla-meta-competitors-task}\\
\toprule\rowcolor{jevvlaheader} \textbf{Task} & \textbf{Direct} & \textbf{JevVLA} & \textbf{Probe} & \textbf{RAS} \\
\midrule\endfirsthead
\toprule\rowcolor{jevvlaheader} \textbf{Task (continued)} & \textbf{Direct} & \textbf{JevVLA} & \textbf{Probe} & \textbf{RAS} \\
\midrule\endhead
\bottomrule\endfoot
assembly & 49 & 49 & 36 & 5 \\
basketball & 14 & 35 & 40 & 12 \\
bin picking & 27 & 22 & 20 & 10 \\
box close & 29 & 30 & 24 & 29 \\
button press topdown & 49 & 49 & 48 & 25 \\
button press topdown wall & 50 & 50 & 50 & 34 \\
button press & 50 & 50 & 50 & 50 \\
button press wall & 50 & 50 & 48 & 50 \\
coffee button & 50 & 50 & 50 & 1 \\
coffee pull & 27 & 20 & 37 & 0 \\
coffee push & 50 & 50 & 49 & 36 \\
dial turn & 18 & 18 & 13 & 0 \\
disassemble & 39 & 39 & 44 & 18 \\
door close & 46 & 47 & 50 & 50 \\
door lock & 45 & 45 & 16 & 50 \\
door open & 50 & 45 & 50 & 50 \\
door unlock & 44 & 50 & 50 & 44 \\
drawer close & 50 & 50 & 50 & 48 \\
drawer open & 50 & 50 & 50 & 30 \\
faucet close & 46 & 49 & 33 & 22 \\
faucet open & 50 & 42 & 38 & 50 \\
hammer & 17 & 24 & 20 & 0 \\
hand insert & 50 & 50 & 41 & 4 \\
handle press side & 41 & 42 & 48 & 39 \\
handle press & 45 & 45 & 48 & 44 \\
handle pull side & 0 & 3 & 0 & 0 \\
handle pull & 7 & 7 & 0 & 7 \\
lever pull & 23 & 23 & 16 & 8 \\
peg insert side & 46 & 50 & 35 & 46 \\
peg unplug side & 46 & 45 & 50 & 9 \\
pick out of hole & 11 & 13 & 10 & 16 \\
pick place & 33 & 32 & 44 & 2 \\
pick place wall & 48 & 49 & 44 & 4 \\
plate slide back side & 50 & 50 & 50 & 2 \\
plate slide back & 48 & 50 & 50 & 18 \\
plate slide side & 50 & 48 & 50 & 44 \\
plate slide & 49 & 50 & 50 & 49 \\
push back & 20 & 23 & 30 & 9 \\
push & 50 & 50 & 50 & 10 \\
push wall & 41 & 41 & 50 & 7 \\
reach & 14 & 14 & 14 & 0 \\
reach wall & 36 & 36 & 39 & 16 \\
shelf place & 29 & 31 & 26 & 20 \\
soccer & 34 & 37 & 11 & 29 \\
stick pull & 17 & 10 & 16 & 0 \\
stick push & 25 & 30 & 19 & 0 \\
sweep into & 47 & 47 & 40 & 50 \\
sweep & 38 & 40 & 42 & 25 \\
window close & 49 & 49 & 50 & 47 \\
window open & 49 & 49 & 50 & 49 \\
\midrule\rowcolor{jevvlatotal} \textbf{Total} & \textbf{1896} & \textbf{1928} & \textbf{1839} & \textbf{1168} \\
\end{longtable}\endgroup
Each entry is the number of successful episodes out of 50 for that task.

\begingroup\footnotesize
\setlength{\tabcolsep}{4pt}\renewcommand{\arraystretch}{1.10}
\begin{longtable}{@{}p{0.43\linewidth}rrrrr@{}}
\caption{CLIPort: online and learned comparison methods by task.}\label{tab:jevvla-cliport-competitors-task-a}\\
\toprule\rowcolor{jevvlaheader} \textbf{Task} & \textbf{Direct} & \textbf{JevVLA} & \textbf{UA} & \textbf{Logistic} & \textbf{MLP} \\
\midrule\endfirsthead
\toprule\rowcolor{jevvlaheader} \textbf{Task (continued)} & \textbf{Direct} & \textbf{JevVLA} & \textbf{UA} & \textbf{Logistic} & \textbf{MLP} \\
\midrule\endhead
\bottomrule\endfoot
align rope & 9 & 6 & 8 & 11 & 8 \\
assembling kits seq seen colors & 22 & 21 & 21 & 20 & 18 \\
assembling kits seq unseen colors & 2 & 0 & 5 & 2 & 0 \\
packing boxes pairs seen colors & 52 & 55 & 57 & 56 & 54 \\
packing boxes pairs unseen colors & 14 & 16 & 30 & 14 & 14 \\
packing seen google objects group & 21 & 25 & 13 & 22 & 25 \\
packing seen google objects seq & 12 & 18 & 26 & 13 & 16 \\
packing shapes & 32 & 34 & 35 & 27 & 28 \\
packing unseen google objects group & 25 & 28 & 28 & 31 & 33 \\
packing unseen google objects seq & 15 & 13 & 8 & 12 & 11 \\
put block in bowl seen colors & 99 & 98 & 97 & 99 & 99 \\
put block in bowl unseen colors & 36 & 45 & 37 & 44 & 38 \\
separating piles seen colors & 65 & 66 & 44 & 64 & 67 \\
separating piles unseen colors & 42 & 30 & 30 & 45 & 43 \\
stack block pyramid seq seen colors & 85 & 86 & 83 & 83 & 85 \\
stack block pyramid seq unseen colors & 0 & 0 & 0 & 1 & 0 \\
towers of hanoi seq seen colors & 63 & 73 & 72 & 62 & 66 \\
towers of hanoi seq unseen colors & 22 & 25 & 13 & 17 & 12 \\
\midrule\rowcolor{jevvlatotal} \textbf{Total} & \textbf{616} & \textbf{639} & \textbf{607} & \textbf{623} & \textbf{617} \\
\end{longtable}\endgroup
Each entry is the number of successful episodes out of 100 for that task.

\begingroup\footnotesize
\setlength{\tabcolsep}{4pt}\renewcommand{\arraystretch}{1.10}
\begin{longtable}{@{}p{0.43\linewidth}rrrrr@{}}
\caption{CLIPort: tabular model architectures by task.}\label{tab:jevvla-cliport-competitors-task-b}\\
\toprule\rowcolor{jevvlaheader} \textbf{Task} & \textbf{Direct} & \textbf{JevVLA} & \textbf{TabM} & \textbf{FT-Trans.} & \textbf{TabNet} \\
\midrule\endfirsthead
\toprule\rowcolor{jevvlaheader} \textbf{Task (continued)} & \textbf{Direct} & \textbf{JevVLA} & \textbf{TabM} & \textbf{FT-Trans.} & \textbf{TabNet} \\
\midrule\endhead
\bottomrule\endfoot
align rope & 9 & 6 & 5 & 5 & 4 \\
assembling kits seq seen colors & 22 & 21 & 23 & 15 & 17 \\
assembling kits seq unseen colors & 2 & 0 & 2 & 2 & 0 \\
packing boxes pairs seen colors & 52 & 55 & 56 & 59 & 54 \\
packing boxes pairs unseen colors & 14 & 16 & 14 & 12 & 15 \\
packing seen google objects group & 21 & 25 & 23 & 28 & 21 \\
packing seen google objects seq & 12 & 18 & 14 & 14 & 22 \\
packing shapes & 32 & 34 & 29 & 30 & 34 \\
packing unseen google objects group & 25 & 28 & 29 & 34 & 25 \\
packing unseen google objects seq & 15 & 13 & 16 & 11 & 9 \\
put block in bowl seen colors & 99 & 98 & 100 & 97 & 98 \\
put block in bowl unseen colors & 36 & 45 & 42 & 42 & 42 \\
separating piles seen colors & 65 & 66 & 64 & 63 & 65 \\
separating piles unseen colors & 42 & 30 & 42 & 44 & 48 \\
stack block pyramid seq seen colors & 85 & 86 & 85 & 82 & 80 \\
stack block pyramid seq unseen colors & 0 & 0 & 2 & 2 & 0 \\
towers of hanoi seq seen colors & 63 & 73 & 67 & 69 & 68 \\
towers of hanoi seq unseen colors & 22 & 25 & 14 & 15 & 12 \\
\midrule\rowcolor{jevvlatotal} \textbf{Total} & \textbf{616} & \textbf{639} & \textbf{627} & \textbf{624} & \textbf{614} \\
\end{longtable}\endgroup
Each entry is the number of successful episodes out of 100 for that task.

\end{document}